\documentclass[12pt]{article}

\usepackage[utf8]{inputenc}
\usepackage[T1]{fontenc}
\usepackage[USenglish]{babel}
\usepackage[top=2.5cm,bottom=3cm,left=2.2cm,right=2.2cm]{geometry}
\usepackage{setspace}
\usepackage{lineno}
\usepackage{graphicx}
\graphicspath{{Figures/}}
\usepackage{booktabs}
\usepackage{multirow}
\usepackage{adjustbox}
\usepackage{array}
\newcolumntype{L}[1]{>{\raggedright\let\newline\\\arraybackslash\hspace{0pt}}m{#1}}
\newcolumntype{C}[1]{>{\centering\let\newline\\\arraybackslash\hspace{0pt}}m{#1}}
\newcolumntype{R}[1]{>{\raggedleft\let\newline\\\arraybackslash\hspace{0pt}}m{#1}}
\usepackage{amsmath,amssymb}
\usepackage{enumitem}
\usepackage{caption}
\usepackage{subcaption}
\usepackage{float}
\usepackage[flushleft]{threeparttable}
\usepackage{gensymb}
\usepackage{pdflscape}
\usepackage{siunitx}
\usepackage{tabularx,ragged2e}
\usepackage{diagbox}
\usepackage{authblk}
\usepackage[numbers,sort&compress,super]{natbib}
\usepackage[colorlinks=true,linkcolor=blue,citecolor=blue,urlcolor=blue]{hyperref}
\usepackage{doi}
\usepackage{orcidlink}

\emergencystretch=\maxdimen
\title{\textbf{Wall-to-wall maps of canopy height and aboveground carbon across the Atlantic Forest}}

\author[1,3,4*]{Matheus Pinheiro Ferreira\,\orcidlink{0000-0003-0687-4422}}
\author[2]{Thiago Carmuega Rabaçal}
\author[3]{Giulio Brossi Santoro\,\orcidlink{0000-0002-5596-5817}}
\author[3]{Giovanna de Andrade Ferreira\,\orcidlink{0000-0001-7714-8327}}
\author[1]{Rodrigo Costa Pinto\,\orcidlink{0000-0002-7613-2590}}
\author[3]{Danilo R. A. de Almeida\,\orcidlink{0000-0002-8747-0085}}
\author[3,4,5]{Paulo Guilherme Molin\,\orcidlink{0000-0002-4587-935X}}
\author[6]{Catherine Torres de Almeida\,\orcidlink{0000-0002-8140-2903}}
\author[5]{Angélica F. Resende\,\orcidlink{0000-0002-9875-1122}}
\author[8]{Vinicius Castro Souza\,\orcidlink{0000-0002-3733-7892}}
\author[8]{Marcelo Antonio Pinho Ferreira\,\orcidlink{0009-0000-4080-2153}}
\author[8]{Cássio Toledo\,\orcidlink{0000-0002-9485-8694}}
\author[9]{Eben North Broadbent\,\orcidlink{0000-0002-4488-4237}}
\author[9]{Angelica Almeyda Zambrano\,\orcidlink{0000-0001-5081-9936}}
\author[10]{Laura Simões\,\orcidlink{0000-0003-1423-3094}}
\author[1,4]{Juliano van Melis\,\orcidlink{0000-0002-2911-3733}}
\author[7,8]{Ricardo Ribeiro Rodrigues\,\orcidlink{0000-0003-4818-0736}}
\author[1,4,7]{Pedro H. S. Brancalion\,\orcidlink{0000-0001-8245-4062}}

\affil[1]{Department of Forest Sciences, ``Luiz de Queiroz'' College of Agriculture, University of S\~ao Paulo (USP/ESALQ), Piracicaba, SP, Brazil}
\affil[2]{Instituto Geográfico e Cartográfico de São Paulo (IGC), São Paulo, SP, Brazil}
\affil[3]{Bioflore Technology, Brazil}
\affil[4]{Center for Carbon Research in Tropical Agriculture (CCARBON), University of São Paulo (USP), Piracicaba-SP, Brazil}
\affil[5]{Center for Nature Sciences, Federal University of S\~ao Carlos (UFSCar), Buri, SP, Brazil}
\affil[6]{Department of Geography, Federal University of Paraná (UFPR), Curitiba, PR, Brazil}
\affil[7]{Re.green, Praça Santos Dumont, 70 – 4° Andar, Rio de Janeiro-RJ, Brazil}
\affil[8]{Department of Biological Sciences, ``Luiz de Queiroz'' College of Agriculture, University of S\~ao Paulo (USP/ESALQ), Piracicaba, SP, Brazil}
\affil[9]{Spatial Ecology and Conservation (SPEC) Lab, School of Forest, Fisheries, and Geomatics Sciences, University of Florida, Gainesville, FL 32611 USA}
\affil[10]{Universidade Federal de S\~ao Carlos, Araras, Brazil}
\affil[*]{Corresponding author. E-mail address: mpferreira@usp.br}

\date{}

\begin{document}
\onehalfspacing
\maketitle

\begin{abstract}
Global canopy height maps are increasingly used to estimate forest carbon, yet their regional accuracy is rarely tested. We present wall-to-wall maps of canopy height (10 m) and aboveground carbon density (30 m) for the Brazilian Atlantic Forest in 2024. Height was predicted by a deep learning network fusing optical (Sentinel-2), SAR (Sentinel-1, ALOS PALSAR) and elevation (ALOS) data, trained and tested on 58,800 km$^2$ of airborne LiDAR from a statewide campaign over São Paulo state, Brazil. The mean absolute error at 10 m is 2.26 m on held-out test patches (1.90 m at 30 m) and 2.8 to 5.5 m at three independent LiDAR sites, against 4.14 m and 3.9 to 13.2 m for the Meta and World Resources Institute global product (version 2) over the same pixels. A power law fitted to 124 inventory plots converted canopy height to above ground carbon density. The remaining forest holds 1,944 TgC over 28.3 Mha. Global products therefore need local validation before use in carbon accounting.
\end{abstract}

\section*{Background \& Summary}

Global and regional canopy height maps derived from satellite imagery have recently been released at resolutions from 30~m down to 1~m \citep{potapov2021mapping,lang2023high,tolan2024very,meta2026chm2, Wagner_etal_2025}. Canopy height maps are increasingly converted into biomass and carbon, either from a global product \citep{becker2025unrealized} or from maps produced for a specific region \citep{schwartz2023forms,liu2023overlooked,wan2025satellite}. Errors in canopy height are therefore carried into the carbon estimates. Yet the accuracy of the global products is rarely known at the regional scale. These products are built to perform everywhere, and independent measurements of canopy height over areas large and varied enough to test them within a region are scarce. Airborne light detection and ranging (LiDAR) surveys that cover entire states or countries can fill this gap. The canopy height model derived from such a survey spans the full range of forest types, relief and land use of a region, and can be used both to evaluate a global product and to train a regional model.

The interest in canopy height stems from its link to biomass. Biomass information is obtained from forest inventories, which measure trees in plots of about 1~ha. Although this is the most accurate method, it covers only small areas and is costly. Remote sensing is the only means of extending plot measurements to entire landscapes, provided that the sensor captures a forest attribute closely related to biomass. Among the forest attributes that can be retrieved from remote sensing, canopy height is the one most closely related to aboveground biomass. At plot scale, top canopy height alone explains most of the variance in biomass density \citep{asner2014mapping}. Canopy height is also the attribute that remote sensing measures most accurately, through LiDAR. A LiDAR sensor emits laser pulses and times their return. Part of each pulse is reflected by the canopy top and part by the ground, through gaps in the foliage, and the difference between the two returns gives canopy height \citep{dubayah2020gedi}. Acquiring airborne LiDAR over hundreds of thousands of hectares, however, is expensive, and spaceborne LiDAR samples the surface only in discrete footprints \citep{dubayah2020gedi}. Large-scale estimates therefore rely on optical and radar imagery, which cover the entire surface at regular intervals. These sensors do not measure height directly. Their signals respond to canopy structure only indirectly and saturate at moderate biomass. They can, however, be calibrated against LiDAR where both are available, and the calibrated model can then be applied wherever imagery exists. This is how most large-scale carbon maps have been produced \citep{saatchi2011benchmark,baccini2012estimated,avitabile2016integrated, Wagner_etal_2025}. Predicting canopy height from imagery calibrated with LiDAR is therefore the most direct route to a wall-to-wall carbon map.

The Brazilian Atlantic Forest is one of the regions where accurate canopy height and carbon maps are most needed. It has been reduced to roughly a quarter of its original extent, and what remains is dispersed over tens of thousands of small remnants and regenerating forest patches embedded in a matrix of pasture, cropland and tree plantation \citep{ribeiro2009brazilian,rezende2018hotspot}. It is also the region where the largest restoration commitments in the tropics are being implemented \citep{brancalion2019global,strassburg2019strategic}, and where fragments hold a quarter to a third less biomass than less disturbed reference stands \citep{lima2020erosion}, human disturbance outweighs climate as a driver of carbon stocks \citep{pyles2022human}, and several million hectares have regrown since 1985 \citep{silvajunior2020benchmark,rosan2026forest}. Quantifying these processes, and identifying where restoration would yield the largest carbon gains, requires biomass and carbon maps fine enough to resolve individual fragments.

Where dense airborne LiDAR is available for training, deep learning models reach errors of a few metres, as shown by the country-wide maps of Switzerland from Sentinel-2 \citep{lang2019country} and of France from Sentinel-1, Sentinel-2 and GEDI \citep{schwartz2023forms}. The Atlantic Forest lacked the training data for an equivalent effort until 2023 and 2024, when an airborne LiDAR campaign was flown over the State of S\~ao Paulo, in south-eastern Brazil. The state covers about 248{,}000~km$^2$, an area comparable to that of the United Kingdom, and lies mostly within the Atlantic Forest region. It contains every forest formation of the region, from the dense evergreen forest of the Serra do Mar, the coastal mountain range, to the seasonal semi-deciduous forest of the interior plateau and the mixed Araucaria forest of the highlands, across the full range of relief and fragmentation the region presents. We used the canopy height model derived from this campaign to train a deep learning model that predicts canopy height from satellite imagery, applied this model to map canopy height and, from it, aboveground carbon across the entire Atlantic Forest, and evaluated the resulting map, side by side with a global canopy height product, against three independent airborne LiDAR surveys.

The dataset has two main layers. The first is a canopy height model of the entire Brazilian Atlantic Forest for 2024 at 10~m resolution, predicted by a deep learning model that combines Sentinel-2 optical imagery, Sentinel-1 and ALOS PALSAR radar imagery, and the ALOS elevation data. The second is an aboveground carbon density map at 30~m resolution, derived from the predicted canopy height through a relationship fitted to 124 forest inventory plots with LiDAR-derived canopy height. The two layers are accompanied by the standard error of the carbon density at 30~m, by a summary of the mean carbon density and its uncertainty at 1~km, and by versions of both maps without the forest mask (``Data Records'').

On held-out test patches in S\~ao Paulo, the predicted canopy height has a mean absolute error of 2.26~m at 10~m resolution and 1.90~m at 30~m. Over the same test pixels, the global canopy height product of Meta and the World Resources Institute \citep{meta2026chm2,tolan2024very} has a mean absolute error of 4.14~m at 10~m. The published 10~m map was then compared with three airborne LiDAR surveys that share no pixel with the training or the test patches, in southern Bahia, in Santa Catarina and at Rio Claro, in the interior of S\~ao Paulo, where its mean absolute error is 2.82, 3.48 and 5.47~m and that of the global product, over the same pixels, is 3.92, 4.47 and 13.23~m. At Rio Claro, a mosaic of old eucalypt stands and native vegetation with a mean canopy height of 28~m, the global product underestimates canopy height by 42\%. This contrast indicates that global canopy height products need regional validation against airborne LiDAR before they are used to estimate biomass and carbon. Canopy height explains 76\% of the variance in plot carbon density.

Summed over the ten natural domains of the Atlantic Forest defined by the Brazilian Institute of Geography and Statistics (IBGE) \citep{ibge2025dominios}, the remaining forest covers 28.3~Mha and holds 1{,}944~Tg~C at a mean density of 68.8~Mg~C~ha$^{-1}$. The model was never given this regionalization. Even so, median carbon density across its domains spans almost a factor of five, from 21~Mg~C~ha$^{-1}$ on the north-eastern coast to 99~Mg~C~ha$^{-1}$ in the dense forest of the eastern south-east, following the gradient of forest cover predicted by the region's fragmentation history. The dataset provides a regionally calibrated baseline for carbon accounting, for tracking regrowth and degradation at the fragment scale, and for prioritizing restoration where commitments are largest.

\section{Methods}
\label{sec:methods}

The dataset was produced in five stages. A statewide airborne LiDAR campaign over the State of S\~ao Paulo was processed into 10~m canopy height models and paired with annual composites of Sentinel-2, Sentinel-1, ALOS PALSAR and the ALOS digital surface model on a grid of 2.56~km patches. A four-branch fusion network was trained on these patches and evaluated on held-out patches of the same grid. The trained network was applied to the entire Atlantic Forest region for the year 2024. The predicted height was aggregated to 30~m and converted into aboveground carbon density with a power law calibrated on 124 inventory plots with co-located LiDAR. Finally, the published canopy height was evaluated against three airborne LiDAR surveys that are independent of the training data, a global canopy height product was evaluated over the same pixels at these sites and in the test patches, and the carbon map was summarized by natural domain. The following subsections describe each stage.

\subsection{Study area}
\label{sec:studyarea}
 
\begin{figure}[!t]
\centering
\includegraphics[width=\textwidth]{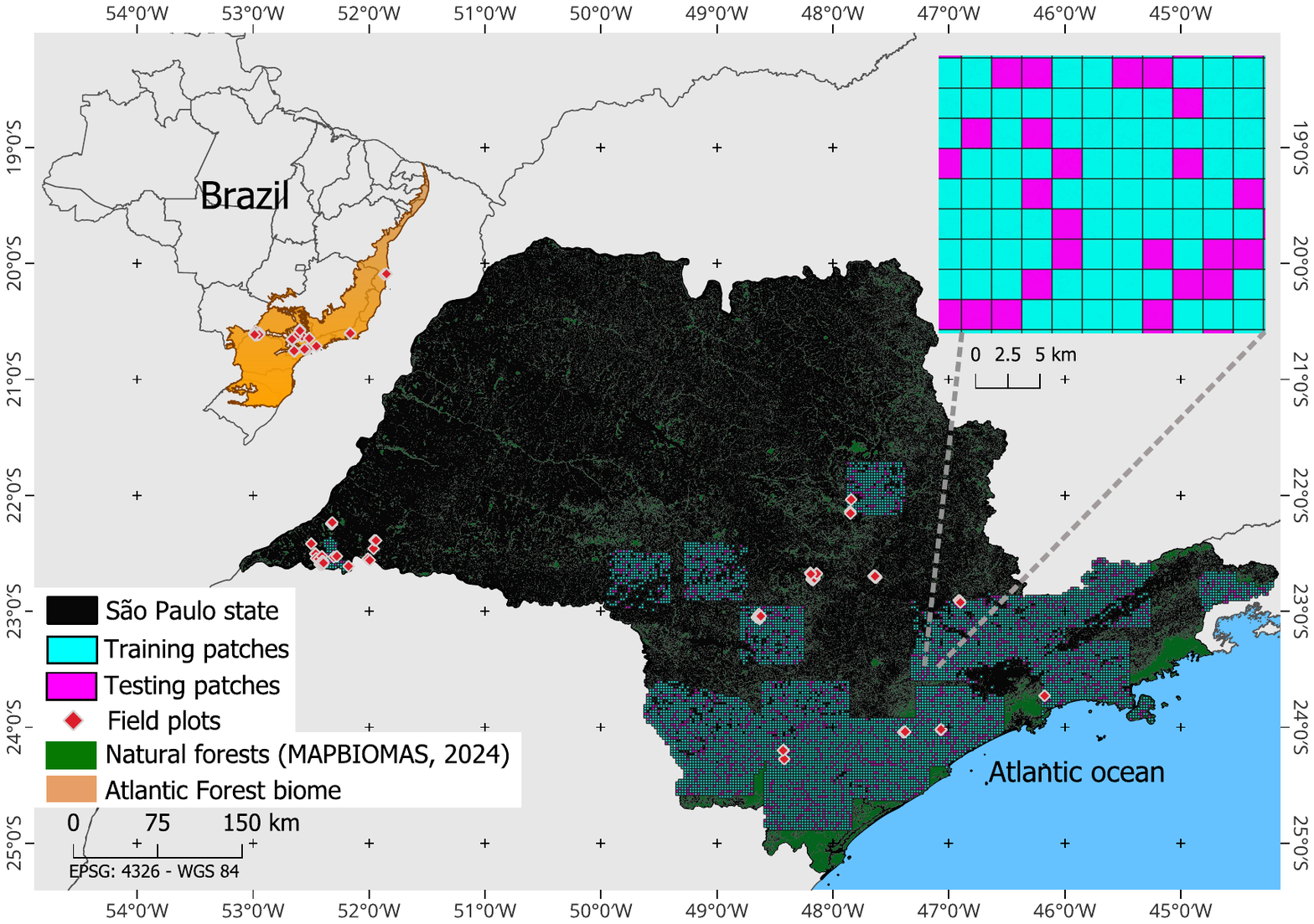}
\caption{Study area. The Atlantic Forest region is outlined in the inset; the main panel shows the State of São Paulo, the reference window from which the training and test patches of the calibration grid were drawn and which contains a variety of forest formations. Cyan and magenta polygons indicate, respectively, the training and testing patches of size 2.56\,$\times$\,2.56~km. Red dots mark the 124 field plots used to convert canopy height into aboveground carbon density.}
\label{Fig:studyarea}
\end{figure}

The study area is the Atlantic Forest region of Brazil (Fig.~\ref{Fig:studyarea}). The Atlantic Forest originally covered some 112~Mha along a wide latitudinal gradient of the Brazilian coast, reaching inland across 15 of the 27 Brazilian states \citep{rezende2018hotspot}, from the tropical north-east to the subtropical south, and from sea level to more than 2{,}000~m in the Serra do Mar and Serra da Mantiqueira ranges. Five centuries of land conversion have reduced it to a fraction of that extent --- estimates range from 11.7--16\% of forest cover \citep{ribeiro2009brazilian} to 28\% of native vegetation when non-forest formations and young regrowth are included \citep{rezende2018hotspot} --- distributed as tens of thousands of small, isolated and strongly edge-affected fragments. Despite this, the region remains a biodiversity hotspot, with high levels of endemism \citep{myers2000biodiversity}, and it is the focus of some of the largest forest restoration commitments in the tropics.
 
\subsubsection{The State of São Paulo as reference window}
\label{sec:studyarea-sp}
 
Along the coastal region, the Atlantic Forest predominates, especially ombrophilous formations characterized by annual rainfall exceeding 2{,}000~mm and a mean annual temperature of approximately 22$^{\circ}$C. These forests, extending across the Serra do Mar mountain range, sustain high levels of endemism and biodiversity \citep{myers2000biodiversity}. The abundant year-round rainfall, without a marked dry season, supports a dense evergreen canopy and the tallest stands in the study area. Inland, the vegetation transitions into seasonal semi-deciduous forests in the central portion of the state, where annual precipitation ranges between 1{,}000 and 1{,}500~mm and the mean annual temperature is about 20$^{\circ}$C. In these forests, a well-defined dry season induces significant leaf shedding \citep{veloso1991classificaccao}. Savanna-like formations of the Cerrado are also present, dominated by a mixture of scattered trees, shrubs and extensive grasslands, occurring where annual rainfall varies from 800 to 1{,}600~mm and mean annual temperature is around 24$^{\circ}$C.
 
This ecological gradient is important to the present study. The seasonal semi-deciduous and savanna formations of the interior undergo a pronounced phenological change between the rainy and the dry season, whereas the coastal ombrophilous forests remain evergreen, so that no single seasonal composite is equally informative everywhere. An annual composite of satellite imagery, which integrates the full phenological cycle, is the acquisition window that behaves consistently across the whole gradient, and it is the one adopted here (``Satellite imagery''). Radar backscatter, which responds to canopy geometry and water content rather than to leaf pigments, is comparatively more stable across seasons and penetrates canopies where optical reflectance saturates, while terrain elevation carries the topographic prior that separates the tall, humid escarpment forests from the shorter, more disturbed plateau remnants. This complementarity is what motivates the multi-sensor design described below.
 
\subsection{LiDAR reference data}
\label{sec:lidar}
 
We used airborne LiDAR data acquired in a statewide campaign covering the entire territory of São Paulo during 2023 and 2024, coordinated by the \textit{Instituto Geográfico e Cartográfico do Estado de São Paulo} (IGC-SP) and conducted with multiple LiDAR sensors. In our dataset the average point density was about 10 points per square metre.

CHM generation was performed with an automated workflow implemented in R with the \texttt{lidR} package \citep{roussel2020lidr}. First, ground returns were identified with the Cloth Simulation Filter (CSF)\citep{zhang2016easy} (\texttt{classify\_ground} with \texttt{csf}). CSF inverts the point cloud and simulates a rigid cloth falling onto it under gravity. The equilibrium position of the cloth approximates the terrain surface from below, and every return within a distance threshold of that surface is classified as ground. We used the package defaults (cloth resolution of 0.5~m, classification threshold of 0.5~m, 500 iterations, a time step of 0.65 and a rigidness of 1) without the slope post-processing option. CSF requires no window-size or slope parameterization, so a single parameter set behaves acceptably over terrain ranging from the flat interior plateau to the escarpments of the Serra do Mar.

Spurious returns were then flagged with the isolated-voxel filter (\texttt{classify\_noise} with \texttt{ivf}). The algorithm voxelizes the point cloud and counts, for every point, the returns present in the $3\times3\times3$ voxel neighbourhood centred on it. A point whose neighbourhood contains fewer than $n$ other returns is labelled as noise (ASPRS class~18) and discarded. We used a voxel side of 10~m and $n=12$, a permissive setting that removes only grossly isolated returns.

Point heights above ground were obtained by triangulation (\texttt{normalize\_height} with \texttt{tin}). Ground returns are used as the vertices of a Delaunay triangulation, the terrain elevation is interpolated linearly inside each triangle at the planimetric position of every return, and this value is subtracted from the return elevation. Normalizing at the exact position of each point avoids the resampling error that an intermediate gridded DTM would introduce. Returns outside the convex hull of the ground points are extrapolated by inverse-distance weighting of their nearest ground neighbours.

CHMs were then produced at 1~m resolution with the point-to-raster algorithm (\texttt{rasterize\_canopy} with \texttt{p2r}), which assigns to each cell the height of the highest return falling inside it, and exported as GeoTIFF. The 1~m CHMs were mosaicked per quadrant, aggregated to the 10~m Sentinel-2 grid by spatial averaging and reprojected to the coordinate reference system of the satellite mosaics (EPSG 4326, WGS 84), so that the reference height and the optical predictors share an identical pixel grid. This grid is regular in geographic coordinates, with a spacing of $8.983\times10^{-5}$ degrees, the angle that corresponds to 10~m at the equator and the convention that Google Earth Engine applies to an export in EPSG 4326 (WGS 84) at a scale of 10~m. The 25 and 30~m grids of the coarser predictors use 2.5 and 3 times this spacing. Pixel sizes in metres are therefore nominal throughout this paper. The size of a pixel along the meridian is within 1\% of its nominal value, whereas its size along the parallel shrinks with the cosine of latitude, to about 9.2~m for a 10~m pixel at the latitude of S\~ao Paulo. In the 10~m CHM, a value of zero marks both bare ground and the absence of LiDAR data, which is why zero heights receive a separate treatment in training (``Training'') and are left out of the accuracy assessment (``Accuracy assessment''). The 10~m CHM is the regression target of all models reported here.

\subsection{Satellite imagery}
\label{sec:imagery}
 
All satellite predictors were assembled in Google Earth Engine (GEE) and exported over the same set of quadrants, so that every experiment shares an identical spatial footprint and differs only in the set of sensors used. We adopted the annual 2024 composite, built as a per-pixel median of all acquisitions between January and December 2024. However, whenever a specific sensor modality lacked coverage for this period, we gap-filled the dataset using the 2023 image acquired closest to the LiDAR flight date. We favoured the annual window approach for two main reasons. First, it is the only time frame that provides gap-free coverage across the entire study area for all sensors, as the ALOS PALSAR mosaic is produced yearly. Second, averaging the full phenological cycle ensures consistent data across the study area's diverse gradient, from evergreen coastal forests to highly seasonal interior ones.
 
Four products enter the network, each at its own native resolution, namely Sentinel-2 (optical, 10~m), Sentinel-1 (C-band SAR, 10~m), ALOS PALSAR (L-band SAR, 25~m), and ALOS DSM (elevation, 30~m). The images of each sensor are read, normalized, and encoded at its native pixel size, and the spatial alignment between modalities is performed inside the network, in feature space (``Multi-sensor fusion network''). This avoids the interpolation artefacts and the artificial sharpening that result from upsampling a coarse product to a 10~m grid before training.
 
\subsubsection{Sentinel-2 (optical)}

We used Sentinel-2 MSI Level-2A bottom-of-atmosphere reflectance (\texttt{COPERNICUS/S2\_SR\_HARMONIZED}). Scenes with more than 50\% cloud cover (\texttt{CLOUDY\_PIXEL\_PERCENTAGE}) were discarded, and the remaining scenes were masked per pixel with the scene-classification (SCL) band, removing cloud shadow, cloud medium probability, cloud high probability, and cirrus (SCL classes 3, 8, 9, and 10) before compositing. Ten spectral bands were retained. The four 10~m bands are B2 (blue, 459--525~nm), B3 (green, 542--578~nm), B4 (red, 649--680~nm) and B8 (NIR, 780--886~nm), and the six 20~m bands are B5 (697--712~nm), B6 (733--748~nm) and B7 (773--793~nm) (red edge), B8A (narrow NIR, 854--875~nm), and B11 (1568--1659~nm) and B12 (2115--2290~nm) (SWIR), the latter six resampled to 10~m with bilinear interpolation. The resulting ten-band stack at 10~m defines the \emph{reference grid}. Reflectance values were kept as scaled integers and normalized at training time by a fixed divisor of 10{,}000, followed by clipping to $[0,1]$. No per-band min--max scaling was applied to the optical data, which keeps the normalization independent of the particular composite being used.

\subsubsection{Sentinel-1 (C-band SAR)}

Sentinel-1 GRD imagery (\texttt{COPERNICUS/S1\_GRD}) in Interferometric Wide swath mode was used with dual polarization (VV and VH), processed with the standard GEE pipeline (thermal-noise removal, radiometric calibration to $\sigma^{0}$ and terrain correction with the SRTM DEM) and expressed in decibels. Per-pixel median composites were built for the same temporal windows and exported at the native 10~m pixel size, which coincides with the optical reference grid.

\subsubsection{ALOS PALSAR (L-band SAR)}

L-band data were taken from the JAXA ALOS/ALOS-2 PALSAR yearly mosaics (\texttt{JAXA/ALOS/PALSAR/YEARLY/SAR}), which provide HH and HV backscatter that has been ortho-rectified and slope-corrected \citep{shimada2014new}. The mosaics were used at their native 25~m pixel size, coarser than the optical grid by a factor of 2.5, and the digital numbers were converted to $\gamma^{0}$ in decibels with the standard calibration,
\begin{equation}
\gamma^{0}_{\mathrm{dB}} = 10\log_{10}\!\left(\mathrm{DN}^{2}\right) - 83 ,
\end{equation}
invalid returns being set to the same $-83$~dB floor that the min--max normalization later maps to zero. L-band backscatter penetrates deeper into the canopy than C-band and saturates at higher biomass levels, which makes it a natural complement to Sentinel-2 in the dense, tall stands of the Atlantic Forest, where optical reflectance saturates \citep{mitchard2009using}. Because the PALSAR product is a single annual mosaic by construction, this sensor exists only in the annual configuration, which is one of the reasons the reported experiments are restricted to the annual composite.

\subsubsection{ALOS DSM (elevation)}
\label{sec:dsm}

Terrain context was introduced through the JAXA ALOS World 3D 30~m digital surface model, version 4.1 (AW3D30 v4.1, \texttt{JAXA/ALOS/AW3D30/V4\_1}), distributed through the GEE catalogue\footnote{\url{https://developers.google.com/earth-engine/datasets/catalog/JAXA_ALOS_AW3D30_V4_1}}. AW3D30 is a global DSM derived by down-sampling the 5~m AW3D product generated from ALOS PRISM stereo triplets acquired between 2006 and 2011, with subsequent versions filling residual voids with auxiliary DEMs and correcting known artefacts \citep{takaku2014generation}. Only the \texttt{DSM} band was used, which stores elevation in metres referenced to the EGM96 geoid. The quality band \texttt{MSK} was not used.

Unlike Sentinel-1, Sentinel-2 and PALSAR, AW3D30 is a \emph{static} product, a global mosaic of adjacent, non-overlapping tiles with no useful time series. It was therefore exported without any date filter and with a mosaic rather than median reducer, at its native 30~m pixel size, over exactly the same quadrants as the other sensors.

The rationale for including elevation is twofold. First, canopy height in the Atlantic Forest region is strongly structured by relief. The tallest remnants concentrate on the steep, humid slopes of the Serra do Mar and Serra da Mantiqueira, while the plateaus carry shorter, more fragmented and more intensively used vegetation, so elevation carries a prior on attainable height that no spectral band conveys. Second, both SAR products are themselves affected by topography, since the local incidence angle modulates backscatter, so an explicit elevation channel lets the fusion module discount the radar branches where relief rather than biomass drives the signal. A third contribution has to be acknowledged. AW3D30 is a surface model derived from optical stereo imagery, so over closed forest it follows the top of the canopy of 2006 to 2011 and not the ground. In a fragmented landscape, the step between a persistent remnant and the surrounding pasture is therefore part of the elevation signal, smoothed by the 30~m sampling, and the branch can draw on this historical canopy signal as well as on the relief. Elevation is also the only modality that is not normalized with statistics estimated from the training set. It is divided by a fixed ceiling of 2{,}100~m (``Input normalization''), safely above the highest value observed in the study area (2{,}043~m, in the Mantiqueira quadrant), so that the scaling of this branch does not change if the training set is extended.

Throughout this paper, the two JAXA products are referred to as ALOS PALSAR (L-band backscatter) and ALOS DSM (elevation), and the abbreviation DSM always denotes the latter.

\subsection{MapBiomas forest mask}
\label{sec:mapbiomas}

We used the MapBiomas Brazil 10~m land use and land cover collection (Collection~11) \citep{souza2020reconstructing} for the year 2024, an annual product classified from Sentinel-2 imagery at 10~m resolution. The classes ``Forest Formation'' (legend code~3), ``Mangrove'' (legend code~5) and ``Wooded Restinga'' (legend code~49) were extracted and binarized into a forest/non-forest mask. The ``Savanna Formation'' class (legend code~4) is not part of the mask, so this step also leaves out the savanna physiognomies of the Cerrado and retains only its forest formations. Because the product is already delivered at 10~m, the mask matches the nominal resolution of the reference grid. It was only aligned to that grid with nearest-neighbour resampling and clipped to the same bounding boxes as the image patches. The mask is stored alongside every patch and plays two roles. It is one of the selection criteria of the patch grid (``Patch extraction''), and it restricts the accuracy assessment to pixels mapped as forest (``Accuracy assessment'').

\subsection{Dataset construction}
\label{sec:dataset}
 
\subsubsection{Patch extraction}
\label{sec:patches}
 
\begin{figure}[!t]
\centering
\includegraphics[width=\textwidth]{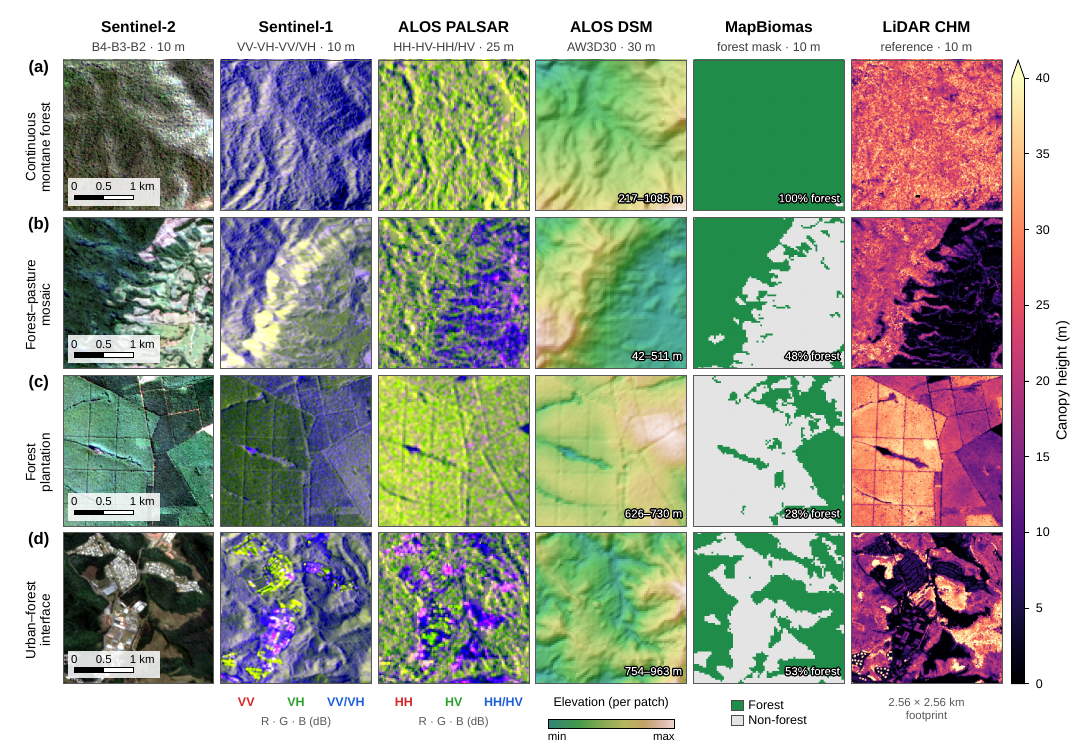}
\caption{Examples of training patches from four contrasting landscapes, (a) continuous montane forest, (b) forest--pasture mosaic, (c) forest plantation and (d) urban--forest interface. Every row covers the same 2{,}560\,$\times$\,2{,}560~m ground footprint, and each source is displayed on its own native grid, without resampling (Table~\ref{tab:patchgeom}). Columns, from left to right, show Sentinel-2 true-colour composite (R: B4, G: B3, B: B2; 10~m); Sentinel-1 false-colour composite (R: VV, G: VH, B: VV/VH ratio, in dB; 10~m); ALOS PALSAR false-colour composite (R: HH, G: HV, B: HH/HV ratio, in dB; 25~m); ALOS DSM rendered as shaded relief (30~m), with the elevation range of the patch in the lower-right corner; MapBiomas 2024 forest mask (``MapBiomas forest mask''), with the percentage of forest pixels in the lower-right corner; and the LiDAR-derived reference CHM (10~m). Sentinel-2 and the DSM are contrast-stretched per patch, whereas each SAR band uses a single 2--98\% percentile stretch shared by the four patches, so that radar tones are comparable across rows. Heights above 40~m are shown with the top colour of the CHM scale. The scale bar in the Sentinel-2 column applies to all panels. In (c), tall planted stands are mapped as non-forest by MapBiomas: they are used in training but excluded from the accuracy assessment (``Accuracy assessment''). In (a) and (b), the slope-driven brightness pattern in both SAR composites illustrates the topographic effect that motivates the DSM branch (``ALOS DSM (elevation)'').}
\label{Fig:patchExamples}
\end{figure}

Patches of 256\,$\times$\,256 pixels at 10~m, a nominal ground footprint of 2{,}560\,$\times$\,2{,}560~m, were laid out on a single \emph{global}, \emph{non-overlapping} grid (Fig. \ref{Fig:studyarea}), anchored at multiples of 256 pixels of the reference grid. Patches were extracted with no lateral overlap. A grid cell was retained as a patch when it met four conditions. At least 10\% of its pixels had a LiDAR height above zero, each of the four predictors covered at least 99\% of its area, at least 10\% of its pixels were mapped as forest in the MapBiomas mask and, for cells within 15~km of the open ocean, no more than 10\% of its pixels were mapped as water. Reference heights above 80~m, which are residual LiDAR noise, were set to no-data. Of the 10{,}472 cells that met the first condition, 8{,}982 met the other three. Ten of them were later discarded because their reference height came from a raster that was not a canopy height model, which left 8{,}972 patches. The ALOS PALSAR and ALOS DSM patches are cut from the upper-left corner of the same footprint. At 25 and 30~m the footprint corresponds to 102.4 and 85.3 pixels, which were rounded to 102 and 85 (Table~\ref{tab:patchgeom}), so these two patches stop 10~m short of the right and lower edges of the footprint, less than half a pixel at their own resolution. Examples of training patches are shown in Fig. \ref{Fig:patchExamples}.

\begin{table}[!t]
\centering
\caption{Geometry of a single training sample. All sources cover the same 2{,}560\,$\times$\,2{,}560~m ground footprint at their native resolution; only Sentinel-2 and Sentinel-1 share the 10~m reference grid.}
\label{tab:patchgeom}
\begin{adjustbox}{max width=\textwidth}
\begin{tabular}{llccc}
\toprule
\textbf{Source} & \textbf{Bands} & \textbf{Resolution (m)} & \textbf{Patch size (px)} & \textbf{Role} \\
\midrule
Sentinel-2   & 10 (B2--B12)  & 10 & 256\,$\times$\,256 & predictor (reference grid) \\
Sentinel-1   & 2 (VV, VH)    & 10 & 256\,$\times$\,256 & predictor \\
ALOS PALSAR  & 2 (HH, HV)    & 25 & 102\,$\times$\,102 & predictor \\
ALOS DSM     & 1 (elevation) & 30 & 85\,$\times$\,85   & predictor \\
LiDAR CHM    & 1             & 10 & 256\,$\times$\,256 & target \\
MapBiomas    & 1 (binary)    & 10 & 256\,$\times$\,256 & selection / evaluation mask \\
\bottomrule
\end{tabular}
\end{adjustbox}
\end{table}
 
\subsubsection{Training and test split}
\label{sec:split}
 
The patch extraction procedure yielded 8{,}972 patches, which together cover about 58{,}800~km$^2$ at the nominal patch size and were divided into 7{,}179 training patches (80.0\%) and 1{,}793 test patches (20.0\%). Because the patches are laid out on a single non-overlapping grid, any two patches are disjoint by construction, and a patch assigned to the test set shares no pixel with any training patch. The split also has to represent the full range of canopy height. The patches are dominated by short canopy, so a plain uniform draw of the same number of patches can, by chance, leave the tall forests that matter most for carbon poorly represented in the test set. We therefore selected the test set by Latin Hypercube Sampling (LHS) on the canopy height of the patches.
 
Let $m_k$ be the median LiDAR canopy height of candidate patch $k$, for $k = 1,\dots,K$, and let $n$ be the number of test patches required, 20\% of $K$. A one-dimensional Latin hypercube design of $n$ points in the unit interval is drawn with \texttt{scipy.stats.qmc.LatinHypercube} and a fixed seed. By construction, the design partitions the interval into $n$ strata of equal probability and places exactly one point in each stratum. Each point is converted into a target height through the empirical quantile function of the $m_k$ and is then matched to the still-unclaimed patch whose median height is closest to that target, so that no patch is selected twice and every design point is realized by an actual patch. The patches so claimed form the test set and the remainder form the training set, which gives the two sets the same distribution of canopy height. The draw was made on the 8{,}982 patches of ``Patch extraction'', before the ten defective patches were removed. It ignores the position of the patches. Test patches are therefore interspersed with training patches over the whole sampled area, 99\% of them share an edge with at least one training patch and no buffer separates the two sets. The test metrics thus measure the error of the model inside the landscapes it was trained on, where neighbouring patches share forest tracts, relief and image acquisitions. The error under extrapolation is assessed separately at the external sites (``External validation sites'').

\subsubsection{Validation split}
\label{sec:validation}
 
From the 7{,}179 training patches, 20\% were held out for validation and early stopping, giving 5{,}743 patches for gradient updates and {1{,}436 for validation. The held-out patches are drawn at random with a fixed seed and are used only to monitor the validation loss and to trigger early stopping. They never contribute to gradient updates. Like the test patches, they are interspersed with the patches used for gradient updates, so the validation error that triggers early stopping is not spatially independent of them. No data augmentation was applied. Given a training set of several thousand non-overlapping patches, we preferred to train on the unmodified samples.
 
\subsubsection{Input normalization}
\label{sec:normalisation}
 
Each modality is normalized to $[0,1]$ with the scheme that suits its physical range, and all constants are stored in the model checkpoint and in a JSON sidecar, then reloaded at inference time so that prediction reproduces exactly the training-time scaling.
 
Sentinel-2 reflectance is divided by the fixed factor of 10{,}000 and clipped to $[0,1]$. Each SAR sensor is normalized independently, band by band, using the absolute minimum $m_b$ and maximum $M_b$ of that band over the patches of that sensor used for gradient updates, so that neither the validation nor the test patches contribute to these statistics,
\begin{equation}
\tilde{x}_{b} = \mathrm{clip}\!\left(\frac{x_{b} - m_{b}}{M_{b} - m_{b}},\, 0,\, 1\right),
\end{equation}
which gave, for the annual composite, $[-66.89,\,33.37]$~dB for Sentinel-1 VV, $[-79.49,\,20.42]$~dB for VH, $[-33.23,\,13.33]$~dB for PALSAR HH and $[-38.44,\,13.06]$~dB for HV.

Separate statistics are mandatory here because L-band annual mosaics and C-band $\sigma^{0}$ in dB have unrelated dynamic ranges and cannot share a normalization. The ALOS DSM branch is the exception: it uses the fixed divisor described in ``ALOS DSM (elevation)'', $\tilde{x} = x/2100$, which is equivalent to $m=0$~m and $M=2{,}100$~m and therefore does not depend on the training set at all. The reference CHM is normalized by a fixed ceiling of $h_{\max}=100$~m.

\subsection{Multi-sensor fusion network}
\label{sec:fusion}

\subsubsection{Overall architecture}

The model (Fig.~\ref{Fig:architecture}) is a single-task dense regression network with $N$ parallel encoder--decoder branches, one per sensor, followed by a cross-modal fusion module and a shared regression head. Formally, given the set of inputs $\{\mathbf{X}_1,\dots,\mathbf{X}_N\}$, with $\mathbf{X}_m \in \mathbb{R}^{C_m \times H_m \times W_m}$, each branch $g_m$ produces a feature map with a common number of channels $C=64$:
\begin{equation}
\mathbf{F}_m = g_m(\mathbf{X}_m) \in \mathbb{R}^{C \times H_m \times W_m}, \qquad m = 1,\dots,N .
\end{equation}
Branch $m=1$ is always Sentinel-2 and defines the reference grid $(H_1,W_1)$. The remaining feature maps are resampled to that grid by bilinear interpolation,
\begin{equation}
\tilde{\mathbf{F}}_m = \mathrm{Interp}_{(H_1,W_1)}(\mathbf{F}_m),\qquad \tilde{\mathbf{F}}_1 = \mathbf{F}_1,
\end{equation}
and the aligned maps are combined by the fusion module $\Phi$ and decoded into a height map by the regression head $\psi$, so that
\begin{equation}
\hat{\mathbf{h}} = h_{\max}\cdot\sigma\!\big(\psi(\Phi(\tilde{\mathbf{F}}_1,\dots,\tilde{\mathbf{F}}_N))\big),
\end{equation}
where $\sigma$ is the logistic function and $h_{\max}$ the height ceiling of ``Input normalization''. The output of the head before the activation, $\psi(\Phi(\cdot))$, is the logit map, and $\ell_{ij}$ denotes its value at pixel $(i,j)$.

Note that the resampling operator $\mathrm{Interp}$ acts on features and never on the input images. The 25~m ALOS PALSAR patch is processed by its own encoder at 25~m, and only its 64-channel semantic representation is brought to the 10~m grid. The same applies to the 30~m DSM branch. This is the mechanism that allows sensors whose pixel sizes differ by a factor of 2.5 or 3 to be combined without degrading the optical detail and without inventing spatial detail in the radar.

\begin{figure}[!t]
\centering
\captionsetup{font=small}

\adjustbox{center}{\includegraphics[width=1.1\textwidth]{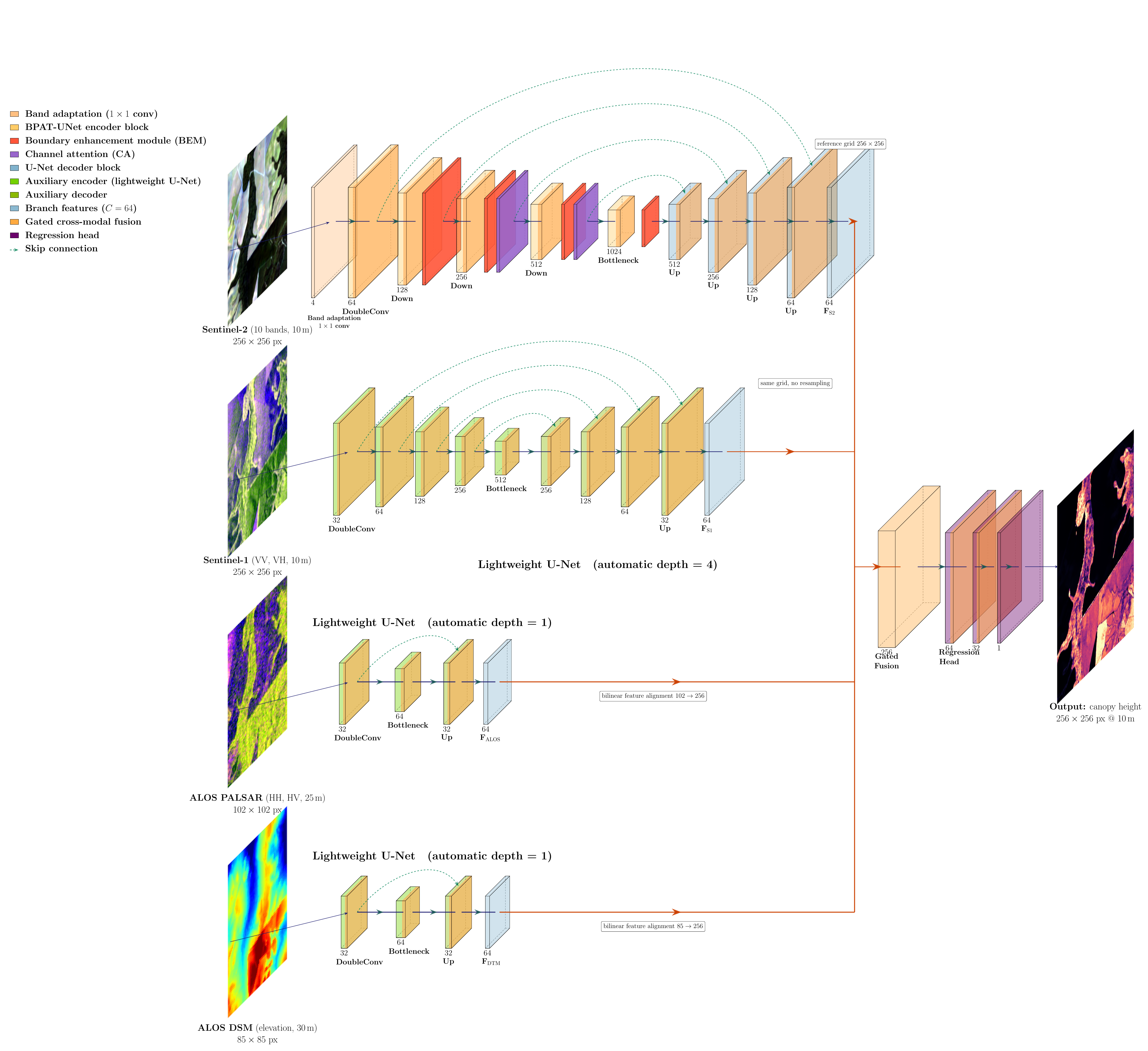}}
\caption{Four-branch fusion architecture for canopy height estimation. Each modality is processed at native resolution: Sentinel-2 via BPAT-UNet (incorporating boundary enhancement [BEM] and channel attention [CA]), while Sentinel-1, ALOS PALSAR, and ALOS DSM use lightweight U-Nets. Branch outputs (64 channels each) are bilinearly aligned to the 256\,$\times$\,256 Sentinel-2 reference grid. A gated cross-modal fusion module combines the 256 channels, and a regression head predicts the final canopy height map. BEM, CA, and fusion blocks are detailed in Fig.~\ref{Fig:architecture-insets}.}
\label{Fig:architecture}
\end{figure}

\subsubsection{Optical branch, the boundary-preserving BPAT-UNet}
\label{sec:bpatunet}
In a fragmented landscape such as the Brazilian Atlantic Forest, the CHM is dominated by abrupt discontinuities, including forest edges against pasture and cropland, roads, and sharp boundaries of \textit{Eucalyptus spp.} plantations. Standard encoder-decoder regressors recover spatial detail only through skip connections and tend to oversmooth these transitions, producing a gradual ramp where the reference LiDAR-derived CHM shows a step. 

We therefore adopted, for the optical branch, an architecture derived from BPAT-UNet \citep{bi2023bpat}, a boundary-preserving U-Net originally proposed for ultrasound segmentation, where the same difficulty arises. The object of interest has a low-contrast but sharp boundary that convolutional decoders blur. We retain the boundary-supervision principle of that architecture in a simplified, fully convolutional form --- the transformer blocks of the original design were replaced by convolutional attention, which is more economical at the patch sizes used here.

The branch begins with a band-adaptation layer, a 1\,$\times$\,1 convolution that projects the ten Sentinel-2 bands into a fixed four-channel representation, so that the same backbone can be instantiated for any number of input bands. The encoder is a sequence of double 3\,$\times$\,3 convolution blocks (each with batch normalization and ReLU) separated by max-pooling, with widths $64\rightarrow128\rightarrow256\rightarrow512\rightarrow1024$; the bottleneck therefore operates at 1/16 of the input resolution (16\,$\times$\,16 pixels for a 256\,$\times$\,256 patch). The decoder mirrors it with transposed convolutions, concatenation of the corresponding encoder feature and a further double convolution, and a final 1\,$\times$\,1 convolution projects the output to the $C=64$ channels that enter the fusion module, or to the single channel of the height logit in the optical baseline (E1), which has no fusion module.

The boundary-preserving behaviour comes from a boundary enhancement module (BEM, Fig. \ref{Fig:architecture-insets}) inserted after each encoder stage. Given a feature tensor $\mathbf{X}\in\mathbb{R}^{C\times H\times W}$, the module estimates a single-channel edge response with a depthwise 3\,$\times$\,3 convolution (one filter per channel, followed by batch normalization and ReLU) and a 1\,$\times$\,1 projection, and converts it into a soft boundary map:
\begin{equation}
\mathbf{B} = \sigma\big(\mathcal{D}(\mathbf{X})\big) \in (0,1)^{1\times H\times W},
\end{equation}
where $\mathcal{D}$ denotes the depthwise edge detector and $\sigma$ the logistic function. In parallel, a standard 3\,$\times$\,3 convolution block produces an enhanced version $\mathcal{E}(\mathbf{X})$ of the same features. The two are combined residually,
\begin{equation}
\mathbf{X}' = \mathbf{X} + \mathcal{E}(\mathbf{X}) \odot \mathbf{B},
\end{equation}
so that the enhancement is applied selectively where an edge is detected and the response is left untouched in homogeneous canopy. The residual formulation means the module can be ignored by the optimizer (by driving $\mathbf{B}$ to zero) if a given stage does not benefit from it, which makes it safe to stack at every scale. Because the BEM acts on the encoder features that are subsequently passed through the skip connections, the sharpened boundary information reaches the decoder at every resolution rather than being reconstructed from the bottleneck alone.

Two of the intermediate stages (256 and 512 channels) additionally carry a channel-attention module, which recalibrates the feature channels using the sum of an average-pooled and a max-pooled descriptor passed through a shared two-layer bottleneck MLP with reduction ratio 16, followed by a logistic gate. Unlike the BEM, which is spatial and edge-driven, this module is purely channel-wise and suppresses the spectral responses that are uninformative for height at that scale. Table~\ref{tab:bpatunet} lists the components and channel dimensions of the branch.

\begin{table}[!t]
\centering
\caption{Components and channel dimensions of the optical (BPAT-UNet) branch, for a 256\,$\times$\,256 Sentinel-2 patch with ten bands. BEM denotes the boundary enhancement module and CA the channel attention module. In the optical baseline (E1), the output projection has a single output channel.}
\label{tab:bpatunet}
\begin{adjustbox}{max width=\textwidth}
\begin{tabular}{lllcc}
\toprule
\textbf{Part} & \textbf{Block} & \textbf{Attention} & \textbf{In $\rightarrow$ Out channels} & \textbf{Spatial size} \\
\midrule
Input     & Band adaptation ($1\times1$ conv)      & --        & $10 \rightarrow 4$      & $256^2$ \\
\midrule
\multirow{5}{*}{Encoder}
 & DoubleConv                                     & --        & $4 \rightarrow 64$      & $256^2$ \\
 & MaxPool + DoubleConv                           & BEM       & $64 \rightarrow 128$    & $128^2$ \\
 & MaxPool + DoubleConv                           & BEM + CA  & $128 \rightarrow 256$   & $64^2$  \\
 & MaxPool + DoubleConv                           & BEM + CA  & $256 \rightarrow 512$   & $32^2$  \\
 & MaxPool + DoubleConv (bottleneck)              & BEM       & $512 \rightarrow 1024$  & $16^2$  \\
\midrule
\multirow{5}{*}{Decoder}
 & Up + Concat (skip, stage 4) + DoubleConv       & --        & $1024+512 \rightarrow 512$ & $32^2$  \\
 & Up + Concat (skip, stage 3) + DoubleConv       & --        & $512+256 \rightarrow 256$  & $64^2$  \\
 & Up + Concat (skip, stage 2) + DoubleConv       & --        & $256+128 \rightarrow 128$  & $128^2$ \\
 & Up + Concat (skip, stage 1) + DoubleConv       & --        & $128+64 \rightarrow 64$    & $256^2$ \\
 & Output projection ($1\times1$ conv)            & --        & $64 \rightarrow 64$        & $256^2$ \\
\bottomrule
\end{tabular}
\end{adjustbox}
\end{table}

\subsubsection{Auxiliary branches}

The Sentinel-1, ALOS PALSAR and ALOS DSM branches are lightweight U-Nets (double 3\,$\times$\,3 convolution blocks with batch normalization and ReLU, max-pooling for downsampling and transposed convolutions with skip connections for upsampling), with a base width of 32 channels and a final 1\,$\times$\,1 convolution projecting to the same $C=64$ channels. Their depth is derived from the native patch side so that the bottleneck never degenerates. A downsampling stage is added while the current side is even and would remain at or above four pixels after halving, up to a maximum of four stages and with a minimum of one. This yields four stages for the Sentinel-1 branch (256~px, bottleneck of 16~px), one stage for the ALOS PALSAR branch (102~px, bottleneck of 51~px, where the rule stops because 51 is odd) and one stage for the ALOS DSM branch (85~px, bottleneck of 42~px, the minimum of one stage, since 85 is already odd). In the decoder, the upsampled tensor is padded to the size of the corresponding skip feature before concatenation, which restores the 85~px side of the DSM branch after its bottleneck is upsampled to 84~px. The shallow depth of the two coarser branches matches the small number of native pixels they carry, about 10{,}400 for ALOS PALSAR and 7{,}200 for the ALOS DSM, against 65{,}536 for the 10~m sensors.

\subsubsection{Gated cross-modal fusion}
\label{sec:gated}

Simple concatenation treats all modalities as equally informative everywhere, which is a poor assumption here. The relative value of L-band backscatter is high where the optical signal saturates, in tall closed-canopy stands, and low over sparse or recently disturbed forests, where Sentinel-2 is already discriminative. We therefore adopted a gated fusion mechanism, in which a global descriptor of all modalities modulates the contribution of each of them, channel by channel (Fig. \ref{Fig:architecture-insets}).

\begin{figure}[!t]
\centering
\captionsetup{font=small}
\adjustbox{center}{\includegraphics[width=1.1\textwidth]{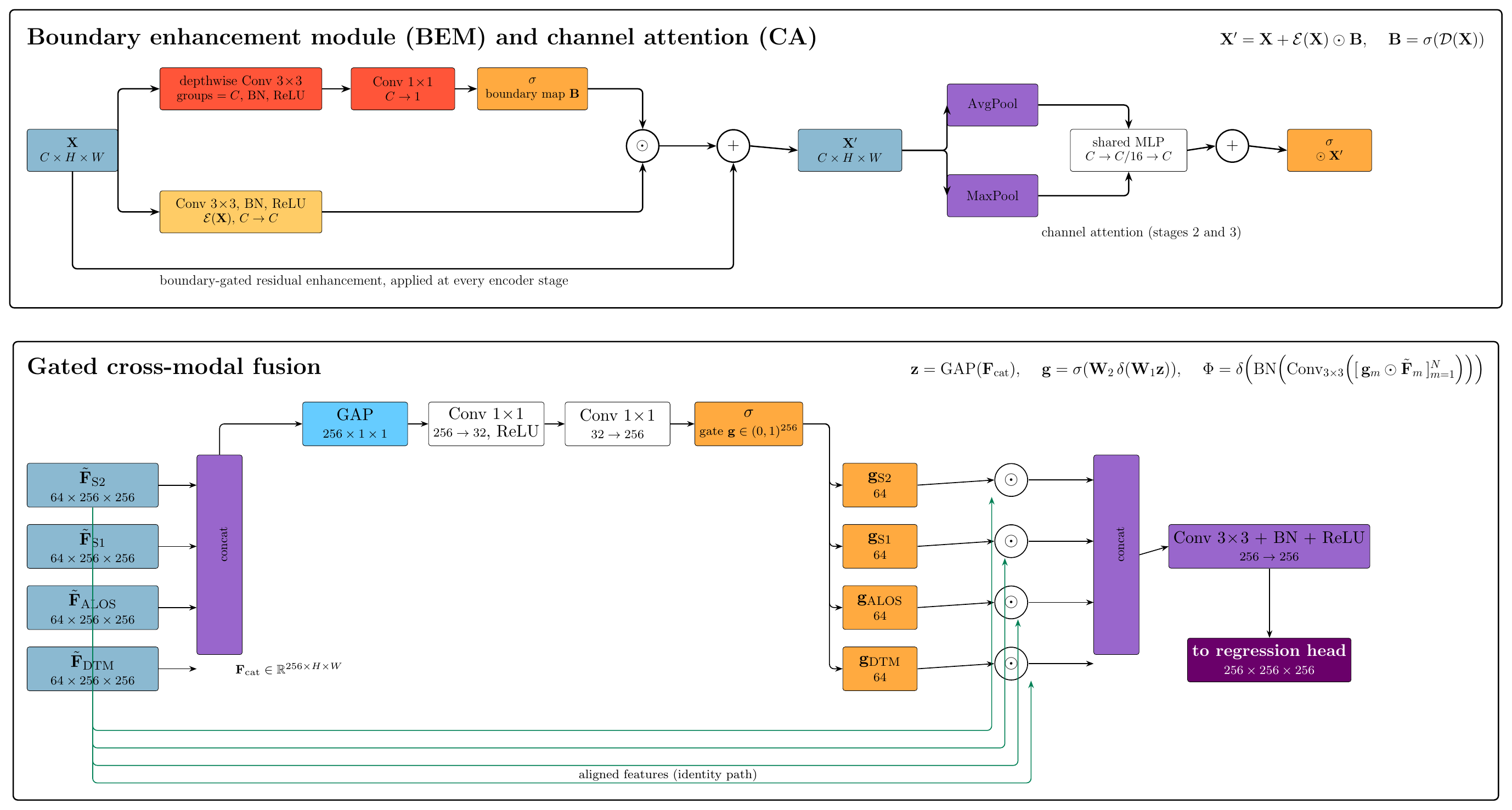}}
\caption{Building blocks of the fusion network (Fig.~\ref{Fig:architecture}). \textbf{Top.} BPAT-UNet's boundary enhancement module (BEM) and channel attention (CA). BEM extracts a single-channel boundary map $\mathbf{B}$ to gate a residual enhancement $\mathcal{E}(\mathbf{X})$ added to the input. CA then reweights the channels of $\mathbf{X}'$ using pooled descriptors and a shared MLP. \textbf{Bottom.} Gated cross-modal fusion. Concatenated features $\mathbf{F}_{\mathrm{cat}}$ are globally pooled and fed to a bottleneck MLP to generate a 64-channel gate per modality. Modality features are channel-wise scaled by their respective gates, re-concatenated, and refined via $3\times3$ convolution before the regression head.}
\label{Fig:architecture-insets}
\end{figure}

Let $\mathbf{F}_{\mathrm{cat}} = [\tilde{\mathbf{F}}_1;\dots;\tilde{\mathbf{F}}_N] \in \mathbb{R}^{NC\times H\times W}$ be the concatenation of the aligned branch features. A global average pooling summarizes the joint state of the sensors into a single vector,
\begin{equation}
\mathbf{z} = \frac{1}{HW}\sum_{i=1}^{H}\sum_{j=1}^{W}\mathbf{F}_{\mathrm{cat}}(:,i,j) \in \mathbb{R}^{NC},
\end{equation}
which conditions a bottleneck multilayer perceptron implemented as two 1\,$\times$\,1 convolutions,
\begin{equation}
\mathbf{g} = \sigma\!\big(\mathbf{W}_2\,\delta(\mathbf{W}_1 \mathbf{z})\big) \in (0,1)^{NC},
\qquad
\mathbf{W}_1 \in \mathbb{R}^{\frac{NC}{r}\times NC},\;
\mathbf{W}_2 \in \mathbb{R}^{NC\times \frac{NC}{r}},
\end{equation}
where $\delta$ is the ReLU non-linearity, $\sigma$ the logistic function and $r=8$ the reduction ratio (with $N=4$ and $C=64$, as in the four-branch model, the hidden layer has 32 units, against 24 in the three-branch configurations and 16 in the two-branch one). The gate vector is partitioned into $N$ blocks of $C$ entries, $\mathbf{g}=[\mathbf{g}_1;\dots;\mathbf{g}_N]$, one per modality, and each branch is re-weighted by its own block before a final concatenation and a 3\,$\times$\,3 convolutional refinement:
\begin{equation}
\Phi(\tilde{\mathbf{F}}_1,\dots,\tilde{\mathbf{F}}_N) \;=\; \delta\Big(\mathrm{BN}\big(\mathrm{Conv}_{3\times3}\big([\,\mathbf{g}_1\odot\tilde{\mathbf{F}}_1;\;\dots\;;\mathbf{g}_N\odot\tilde{\mathbf{F}}_N\,]\big)\big)\Big),
\end{equation}
with $\odot$ denoting channel-wise multiplication broadcast over space. Because $\mathbf{z}$ pools all modalities, the weight assigned to a radar channel depends on what the optical branch is seeing in the same patch, which is the behaviour required when optical saturation is the limiting factor. The fusion module preserves the channel count of the concatenation ($NC$ channels) and therefore leaves the interface with the regression head unchanged; the mean value of each gate block was logged during training as a diagnostic of how much each sensor is effectively contributing.

\subsubsection{Regression head}

The fused tensor of $NC$ channels is decoded by a head composed of a 3\,$\times$\,3 convolution to 64 channels, a 3\,$\times$\,3 convolution to 32 channels --- both followed by batch normalization and ReLU --- and a final 1\,$\times$\,1 convolution to a single channel. A logistic activation is applied outside the network and the result is rescaled by $h_{\max}=100$~m, so that predictions are bounded to a physically plausible range by construction. The output map has the same size as the Sentinel-2 patch (256\,$\times$\,256 at 10~m).

\subsection{Training}
\label{sec:training}

All models were implemented in PyTorch and trained end-to-end from random initialization on a single GPU (NVIDIA GeForce RTX 5090 with 32 GB of VRAM). No branch was pre-trained. The ten-band Sentinel-2 input and the SAR inputs have no counterpart in natural-image datasets, and the boundary-preserving backbone described in ``Optical branch, the boundary-preserving BPAT-UNet'' has no ImageNet checkpoint, so transfer learning was not applicable. The loss is the mean absolute error between the predicted and the reference height in normalized units,
\begin{equation}
\mathcal{L} = \frac{1}{|\Omega|}\sum_{(i,j)\in\Omega}\left| \sigma(\ell_{ij}) - \frac{h_{ij}}{h_{\max}} \right|,
\end{equation}
where $\ell_{ij}$ is the logit of pixel $(i,j)$ (``Overall architecture''), $h_{ij}$ its reference height and $\Omega$ the set of pixels of the patch that enter the loss. Because a height of zero also marks the absence of LiDAR data (``LiDAR reference data''), pixels with a reference height of exactly zero are excluded from $\Omega$ in the patches where they make up at least 5\% of the pixels, where they are typically gaps in the LiDAR coverage, and are kept in the other patches, where they are typically open ground. The same rule applies to the validation error. Pixels outside the forest mask, such as pastures, crops, tree plantations and urban areas, are not excluded, so the network is trained on every land cover, whereas the accuracy assessment is restricted to forested areas (``Accuracy assessment''). MAE was preferred over MSE because of its lower sensitivity to the extreme heights that occasionally survive the LiDAR noise filtering, and because it is directly interpretable in metres after rescaling.

Optimization used AdamW \citep{zhou2024towards} with a base learning rate of $10^{-3}$ and a weight decay of $10^{-6}$, a batch size of 15 patches, and a schedule composed of a 5-epoch linear warm-up from zero followed by a cosine decay that reaches a floor of $10^{-4}$ at epoch 60 and stays there. Training ran for a maximum of 1{,}000 epochs with early stopping on the validation MAE (patience of 25 epochs, minimum improvement of $10^{-5}$), and the checkpoint with the lowest validation MAE was retained. The same hyper-parameters and the same random seed were used in all experiments, so that the comparison between sensor configurations is not confounded by different parameter tuning.

\subsection{Experimental design}
\label{sec:experiments}

We compared five sensor configurations, all trained on the annual 2024 composites (Table~\ref{tab:experiments}). The series isolates, in turn, the contribution of Sentinel-1, of ALOS PALSAR and of the ALOS DSM, always on top of the same optical baseline. When a branch is switched off, the network is instantiated with fewer branches and the gated fusion module is resized accordingly. The optical baseline (E1, $N=1$) has neither fusion module nor regression head, and its BPAT-UNet maps the Sentinel-2 patch directly to the height logit. Everything else is held constant, including the patches, the split, the normalization strategy and the hyper-parameters.

\begin{table}[H]
\centering
\caption{Experimental design. Five sensor configurations, all trained on the annual 2024 Sentinel-2 composite. In the fusion configurations (E2 to E5), every branch contributes 64 feature maps to the gated cross-modal fusion.}
\label{tab:experiments}
\begin{adjustbox}{max width=\textwidth}
\begin{tabular}{llc}
\toprule
\textbf{ID} & \textbf{Configuration} & \textbf{Branches} \\
\midrule
E1 & S2                          & 1 \\
E2 & S2 + S1                     & 2 \\
E3 & S2 + S1 + ALOS PALSAR       & 3 \\
E4 & S2 + S1 + DSM               & 3 \\
E5 & S2 + S1 + ALOS PALSAR + DSM & 4 \\
\bottomrule
\end{tabular}
\end{adjustbox}
\end{table}

The four-branch configuration (E5) is the one that produced the dataset. Like every other configuration, it was trained only on the S\~ao Paulo patches of ``Training and test split'', so the ablation isolates the contribution of each sensor over the whole series, and the three surveys of ``External validation sites'' remain independent of every patch the model saw.

The choice of the optical branch was assessed on the optical baseline, the only configuration in which the whole network is the architecture under test. In addition to the BPAT-UNet, E1 was trained with two encoder--decoder networks of the \texttt{segmentation-models-pytorch} library, a U-Net \citep{ronneberger2015u} and an MA-Net \citep{fan2020ma}, both with a ResNet-50 encoder \citep{he2016deep} initialized with ImageNet weights \citep{deng2009imagenet} and truncated after four downsampling stages, so that, as in the BPAT-UNet, the bottleneck operates at 1/16 of the input resolution, and with decoders of 512, 256, 128 and 64 channels. The three networks were trained anew for this comparison with the same patches, split, normalization, loss, optimizer and learning-rate schedule, and with an early-stopping patience of 10 epochs instead of 25, and were evaluated as in ``Accuracy assessment''. Only the BPAT-UNet starts from random initialization, so the comparison gives the two reference networks the advantage of pre-training.

\subsection{Inference}
\label{sec:inference}

Wall-to-wall prediction was performed tile by tile, on 762 tiles of 5{,}120\,$\times$\,5{,}120 pixels at 10~m. The network was not applied to a whole tile at once. A window of 256\,$\times$\,256 pixels, the size used in training, was moved over each Sentinel-2 tile with an overlap of 50\% between consecutive windows, so that the global descriptor of the gated fusion (``Gated cross-modal fusion'') is always computed over the footprint it saw in training. For each window, the Sentinel-1 window was cut at the same position, and the ALOS PALSAR and ALOS DSM windows of 102 and 85 pixels were cut on their own grids from the position that corresponds to the upper-left corner of the Sentinel-2 window, as in training. All inputs were normalized with the constants stored in the checkpoint, and the logistic activation and the rescaling by $h_{\max}$ were applied to the logits. The overlapping predictions were then averaged with the weights of a two-dimensional Hann window, which gives little weight to the border of each window and avoids seams between windows. No margin was predicted around the tiles, so a pixel at the edge of a tile has image context on one side only. The resulting height map was written as a GeoTIFF that inherits the geotransform and the coordinate reference system of the input tile. The forest mask was applied to the prediction at 10~m. The masked height was then aggregated to 30~m by spatial averaging, on a grid anchored globally so that the cells coincide between tiles, and Equation~\ref{eq:agcd} was applied to the aggregated height. A 30~m pixel receives a value when at least half of the 10~m pixels inside it are forest, and its carbon density then refers to its forest part.

\subsection{Accuracy assessment}
\label{sec:accuracy}

Model accuracy was assessed on the held-out test patches defined in ``Training and test split'', which share no pixel with the training patches of the S\~ao Paulo grid. For each test patch, the predicted CHM and the LiDAR-derived CHM were intersected pixel by pixel and stacked across all patches into two paired vectors.

Two filters were applied before computing the statistics. First, only pixels mapped as forest in the MapBiomas 2024 mask were retained, that is, pixels classified as \emph{Forest Formation}, \emph{Mangrove} or \emph{Wooded Restinga} (``MapBiomas forest mask''), which removes croplands, pastures, water and built-up areas from the comparison. Second, only pixels with a reference LiDAR height above zero were kept, which discards the cells without LiDAR data, the bare ground and the unvegetated surfaces that survive inside the mask. The metrics therefore describe the error over standing forest canopy and do not include the error at the gaps and bare patches that a height of zero would represent.

From the resulting paired vectors we computed the root mean squared error (RMSE), the mean absolute error (MAE), the bias, the coefficient of determination ($R^2$) of an ordinary least-squares fit of prediction against reference and the concordance correlation coefficient (CCC) \citep{lin1989concordance}, using \texttt{scikit-learn} and \texttt{SciPy}. Bias is defined as the mean of LiDAR minus predicted height, so that positive values denote underestimation, and it is also reported as a percentage of the mean reference height. Agreement was visualized with two-dimensional density plots based on hexagonal binning, with equal axes, the 1 to 1 line and the fitted regression line. Residuals were further stratified into reference-height classes and the bias was reported per class.

\subsection{Aboveground carbon density model}
\label{sec:agcd}

We used forest inventory data from 124 field plots of 900~m$^2$ (30 by 30~m) allocated in restoration plantings, naturally regenerated forests and conserved and degraded forest remnants across the Atlantic Forest. Sixty-one plots are restoration plantings, 26 are naturally regenerated forests, 21 are conserved remnants and 16 are degraded remnants. Most of them lie in the State of S\~ao Paulo (114 plots), with five in Rio de Janeiro and five in southern Bahia, and they cover the three main phytophysiognomies of the region, namely seasonal semi-deciduous forest (50 plots), the transition to savanna (46) and ombrophilous forest (28) (Fig.~\ref{Fig:studyarea}). The field protocol required all woody individuals with a diameter at breast height (DBH) equal to or above 5~cm to be recorded, with measurements of DBH and total height. Botanical samples were collected for each species within each plot and later identified by taxonomic specialists, with voucher specimens deposited at the ESA Herbarium of the University of S\~ao Paulo. All plots were georeferenced with dual-frequency GNSS, which ensures high positional accuracy. The forest inventory data were processed to fix or remove data inconsistencies such as typing errors and misspellings, and aboveground biomass was then estimated at tree and plot level with the allometric equation of \citet{chave2014improved}.

Every plot was covered by airborne laser scanning, 93 of them with remotely piloted aircraft systems and the remaining 31 with airborne surveys. One campaign used the GatorEye Lite system, a customized DJI Matrice 300 RTK carrying a Zenmuse L1 sensor together with a multispectral sensor, between 17 and 27 August 2022, and represents, to our knowledge, the largest collection of LiDAR data from remotely piloted aircraft in the Atlantic Forest. The Zenmuse L1 has a field of view of 70$^\circ$, a maximum range of 190~m at 10\% reflectance and a maximum multiple-return rate of 480{,}000 points per second. Three returns per pulse were recorded with a sampling rate of 240~kHz in non-repetitive scanning mode, and the flights were conducted between 90 and 110~m above ground level, at a ground speed of 8 to 10~m~s$^{-1}$ and with a spacing of 20 to 25~m between flight lines, which gives a side overlap of about 82\% at 100~m above ground level. A second campaign, in 2023, used a DJI Matrice 350 RTK with a Zenmuse L2 sensor, which has a field of view of 70$^\circ$, a maximum range of 250~m at 10\% reflectance and a maximum multiple-return rate of 1{,}200{,}000 points per second. Two returns per pulse were stored, with a sampling rate of 240~kHz in non-repetitive scanning mode, at 100~m above ground level, a ground speed of 10~m~s$^{-1}$ and a side overlap of 50\%. The remaining plots were covered by airborne surveys flown by partner companies with a RIEGL LMS-Q680i full-waveform scanner, with a field of view of 45$^\circ$ and a pulse repetition rate of 400~kHz, at a mean altitude of 620~m, a side overlap of 30\% and a ground speed of 62~m~s$^{-1}$.

The point clouds were processed exactly as the LiDAR reference data of this study (``LiDAR reference data''), so that the height used to calibrate the carbon model and the height the network is trained to predict are defined in the same way. For every plot, we then computed the top canopy height (TCH) as the mean of the 1~m CHM pixels falling inside the plot, which is the plot-level counterpart of the surface the model predicts. Plot biomass was converted into aboveground carbon density (AGCD), in Mg~C~ha$^{-1}$, with a carbon fraction of 0.47 \citep{ipcc2006guidelines}. Because the two sides of the relationship must describe the same forest, every plot was required to have its LiDAR acquisition and its field inventory less than one year apart.

Carbon density was then related to canopy height by the power law
\begin{equation}
\mathrm{AGCD} = a\,\mathrm{TCH}^{\,b}.
\label{eq:agcd}
\end{equation}

Equation~\ref{eq:agcd} is what converts the canopy height predicted by the network into aboveground carbon density. The carbon density of a pixel therefore carries three sources of uncertainty, namely the coefficients of the curve, the scatter of individual stands around it, which includes the allometric error of plot biomass, and the error of the predicted height. The first two are propagated into the uncertainty layers of the dataset and the third is treated separately (``Uncertainty of the carbon density map'').

\subsection{Uncertainty of the carbon density map}
\label{sec:agcd_uncertainty}

The uncertainty of the carbon layer was propagated pixel by pixel from Equation~\ref{eq:agcd} with a first-order Taylor expansion, the delta method \citep{verhoef2012delta}, as in the error propagation of plot-based biomass estimates \citep{chave2004error}. Two sources of error enter it, the uncertainty of the coefficients $a$ and $b$, which is shared by every pixel, and the residual scatter of individual stands around the curve, which contains the structural difference between plot categories and the random part of the allometric error of plot biomass. Because that scatter grows with height, Equation~\ref{eq:agcd} was refitted on the 124 plots with a residual standard deviation that is a power function of height,
\begin{equation}
\mathrm{AGCD}_j = a\,\mathrm{TCH}_j^{\,b} + \varepsilon_j, \qquad \varepsilon_j \sim \mathcal{N}\!\left(0,\,\sigma^2(\mathrm{TCH}_j)\right), \qquad \sigma(\mathrm{TCH}) = s_0\,\mathrm{TCH}^{\,\delta},
\label{eq:agcd_var}
\end{equation}
with $\mathrm{AGCD}_j$ in kg~m$^{-2}$ and $a$, $b$, $s_0$ and $\delta$ estimated jointly by maximum likelihood, as \texttt{gnls} with a \texttt{varPower} variance function does in the \texttt{nlme} package of R \citep{pinheiro2000mixed}. The fit also gives the covariance matrix $\boldsymbol{\Sigma}_{\theta}$ of $\boldsymbol{\theta} = (a, b)^{\top}$. Its coefficients, $a = 0.0898$ and $b = 1.650$, are close to but not identical with the pair adopted to generate the carbon layer, $a = 0.089$ and $b = 1.645$, and over the 124 plots the curve of the product lies on average 2.1\% below the refitted one. The refit supplies only the covariance matrix and the variance function, whereas the carbon density itself and the gradients below use the coefficients of the product.

For a 30~m pixel of height $x_i$ the carbon density is $\hat{y}_i = f(x_i;\boldsymbol{\theta}) = 10\,a\,x_i^{\,b}$, in Mg~C~ha$^{-1}$, and its variance is
\begin{equation}
\mathrm{Var}(\hat{y}_i) \approx \mathbf{g}_i^{\top}\boldsymbol{\Sigma}_{\theta}\,\mathbf{g}_i + 100\,\sigma^2(x_i), \qquad \mathbf{g}_i = \left.\frac{\partial f}{\partial \boldsymbol{\theta}}\right|_{x_i} = 10\left[\,x_i^{\,b},\; a\,x_i^{\,b}\ln x_i\,\right]^{\top},
\label{eq:var_pixel}
\end{equation}
the variance of the mean curve at that height plus the residual variance of a single stand, with the gradient evaluated at $a = 0.089$ and $b = 1.645$. Its square root is the standard error $\mathrm{SE}_i$ of the pixel, distributed as record 3 and mapped in Fig.~\ref{Fig:agcdUncertainty}. Because the height $x_i$ is treated as known, $\mathrm{SE}_i$ excludes the error of the predicted height and is a lower bound.

The same expansion gives the term that the error of the predicted height would add. If $\sigma_{x_i}$ is the standard error of the 30~m height of the pixel, the additional variance is
\begin{equation}
\left(\frac{\partial f}{\partial x_i}\right)^{2}\sigma_{x_i}^{2} = \left(10\,a\,b\,x_i^{\,b-1}\right)^{2}\sigma_{x_i}^{2} = \left(\frac{b\,\hat{y}_i}{x_i}\right)^{2}\sigma_{x_i}^{2}.
\label{eq:var_height}
\end{equation}

With the RMSE of 2.58~m of the 30~m height (``Accuracy across sensor configurations'') taken as $\sigma_{x_i}$, this term amounts to a standard error of about 26~Mg~C~ha$^{-1}$ for a stand of 20~m, of the same order as the 34.4~Mg~C~ha$^{-1}$ that Equation~\ref{eq:var_pixel} gives at that height, and the two combine to about 43~Mg~C~ha$^{-1}$. The term was not included in records 3 and 4 for two reasons. The error of the predicted height is not a single number, since it varies with height and between regions (``Technical Validation''). Part of it is also a systematic underestimation of tall canopy, which does not average out over an area and which the exponent of Equation~\ref{eq:agcd}, being above one, amplifies. Users can add Equation~\ref{eq:var_height} with the height error that suits their region and height class.

An error in $a$ and $b$ shifts all pixels together, so the pixel variances cannot be summed as independent. For the mean $\bar{y}$ of $n$ forest pixels, either a cell of about 1~km (record 4) or the whole Atlantic Forest region,
\begin{equation}
\mathrm{Var}(\bar{y}) = \bar{\mathbf{g}}^{\top}\boldsymbol{\Sigma}_{\theta}\,\bar{\mathbf{g}} + \frac{100}{n^{2}}\sum_{i=1}^{n}\sigma^2(x_i) + \frac{s^{2}}{n},
\label{eq:var_cell}
\end{equation}
where $\bar{\mathbf{g}}$ is the mean gradient of the pixels and $s^{2}$ the variance of $\hat{y}_i$ among them. The first two terms are the model-based variance of the mean of the mapped pixels, conditional on their heights. The third term, $s^{2}/n$, is not an estimation error of that mean, since every forest pixel of the cell is mapped. It expresses the spread of carbon density inside the cell, scaled as the sampling variance of a mean, and it makes the reported standard error conservative. Record 4 stores the three terms in separate bands, so the conditional variance can be recovered by leaving the third one out. The last two terms shrink as $n$ grows, whereas the coefficient term does not.

\subsection{External validation sites}
\label{sec:external}

The S\~ao Paulo patches span the forest formations of the Atlantic Forest region (``The State of S\~ao Paulo as reference window''), but the held-out test patches are interspersed with the training patches and share their landscapes, their relief and their image acquisitions (``Training and test split''), so the test metrics measure the error of the model inside the window in which it was calibrated. Users of the wall-to-wall product need to know how that error behaves where the model was not calibrated. We therefore assessed the published product directly, on three airborne LiDAR surveys that share no pixel with the training or the test patches. Two of them lie outside S\~ao Paulo, in southern Bahia and in Santa Catarina, and the third covers a State Forest in the interior of S\~ao Paulo that is isolated from every patch.

\subsubsection{Serra do Conduru State Park}

The first two surveys come from the Sustainable Landscapes Brazil project (\url{https://www.paisagenslidar.cnptia.embrapa.br/}), which distributes surveys over Brazilian sites at no cost. The first is the survey of the Serra do Conduru State Park, in the municipalities of Uru\c{c}uca, Itacar\'e and Ilh\'eus, in southern Bahia \citep{santos2024conduru}. The forests in these regions provide the clearest example of the stature of native forest that the S\~ao Paulo patches lack, with mature stands storing of the order of 390~Mg~ha$^{-1}$ of aboveground biomass and secondary stands recovering biomass at rates among the highest reported for the Neotropics \citep{becknell2018landscape,rocha2026aboveground}. The flights were carried out between 16 and 23 December 2015 with an Optech Orion M300 scanner at a mean altitude of 850~m and a field of view of 12$^\circ$, which gave a mean return density of 94.5 points per square metre, of which 46.2 are first returns. The point cloud is delivered already classified, in SIRGAS 2000 UTM zone 24S. We derived the CHM with the same procedure used for the calibration data (``LiDAR reference data''), at 1~m resolution, and aggregated it to 10~m by spatial averaging, which yielded 4{,}549~ha of valid canopy height.

\subsubsection{Jaragu\'a do Sul and Massaranduba}

The second is the survey of Jaragu\'a do Sul and Massaranduba, in Santa Catarina \citep{santos2023massaranduba}, about 1{,}000~km to the south of Conduru and close to the southern limit of the dense rainforest. It was flown on 26 August 2017 with an Optech ALTM 3100 scanner at a mean altitude of 850~m, a field of view of 15$^\circ$, a scanning frequency of 40~Hz and 70\% overlap between flight lines, which gave a mean return density of 54.05 points per square metre over twelve blocks covering 510~ha, delivered already classified in SIRGAS 2000 UTM zone 22S. The block sits on the eastern escarpment of the Serra do Mar and its ground elevation ranges from about 70 to 920~m over a horizontal distance of roughly 3~km, a relief gradient steeper than that of almost any calibration patch. The CHM was derived and aggregated exactly as for Conduru, which yielded 516~ha of valid canopy height.

\subsubsection*{Edmundo Navarro de Andrade State Forest}

The Edmundo Navarro de Andrade State Forest (FEENA), known locally as the Horto Florestal de Rio Claro, is a protected area of about 2{,}230~ha in the municipality of Rio Claro, in the interior of the State of S\~ao Paulo \citep{fflorestal2024feena}. It was established in 1909 by the Companhia Paulista de Estradas de Ferro, a railway company, to produce wood for sleepers and charcoal. Over the following decades it became the main centre for research on eucalypt species and provenances in Brazil, and its most important species collection was planted in 1919. The former forest station was designated a State Forest, a sustainable-use protected area, in 2002 (State Decree 46{,}819).

The vegetation of FEENA is dominated by old stands of \textit{Eucalyptus}, planted in blocks and representing about 60 species, some close to a century old \citep{fflorestal2024feena}. Native vegetation occurs in three settings. Riparian remnants of seasonal semi-deciduous forest occupy the margins of the watercourses. A native understorey has established beneath many of the eucalypt stands, with different degrees of regeneration. Regenerating Cerrado vegetation also occurs in some stands where the eucalypts were removed. FEENA borders the urban area of Rio Claro and is the largest wooded area in a landscape where native vegetation persists only as small forest fragments and secondary vegetation \citep{fflorestal2024feena}.

The third survey covers the whole of FEENA. It was flown in 2025 with a DJI Zenmuse L2 scanner on board a DJI Matrice 350 RTK aircraft, at 120~m above the ground and at a speed of 8~m~s$^{-1}$, which gave a mean density of 255 points per square metre. The CHM was produced following the procedure described in “LiDAR reference data” at 1 m resolution, and aggregated it to 10 m by spatial averaging.

FEENA brings to the validation three conditions that the other two sites do not offer. It is isolated from the calibration data, since no training or test patch overlaps the State Forest and the nearest patch of either set lies 23~km away. Its canopy is the tallest of the three sites, with a mean LiDAR height of 28.3~m at 10~m resolution. Finally, it combines forest plantations and native vegetation in a single block, so the product is assessed there over forest plantations as well, a land cover that is not part of the forest mask and therefore of the test metrics.

\subsubsection{Canopy height at the three sites}

The three sites are displaced towards taller canopy than the calibration window, and they are displaced in different ways. Half of the pixels of Conduru stand above 19.2~m and half of those of Jaragu\'a do Sul above 18.4~m, compared with 15.7~m in the S\~ao Paulo test patches. Conduru carries the wider distribution of the two, with a 95th percentile of 27.2~m and a tallest pixel of 43.8~m, while Jaragu\'a do Sul is narrower and shorter at the top, with a 95th percentile of 24.6~m. Rio Claro stands apart from both. Its median is 29.7~m and its 95th percentile 47.9~m, a quarter of its pixels exceed 39.8~m and the tallest one reaches 69.5~m, so that 49.4\% of its pixels lie above 30~m, a height exceeded by 1.4\% of the test pixels of S\~ao Paulo. The three sites therefore take the product away from the conditions under which the model was fitted along different axes, namely stature at Conduru, relief and latitude at Jaragu\'a do Sul, and very tall canopy with a mixture of plantations and native vegetation at Rio Claro. Figure~\ref{Fig:histCHM} compares the four distributions over the pixels that enter the respective assessments.

\begin{figure}[!t]
\centering
\includegraphics[width=\textwidth]{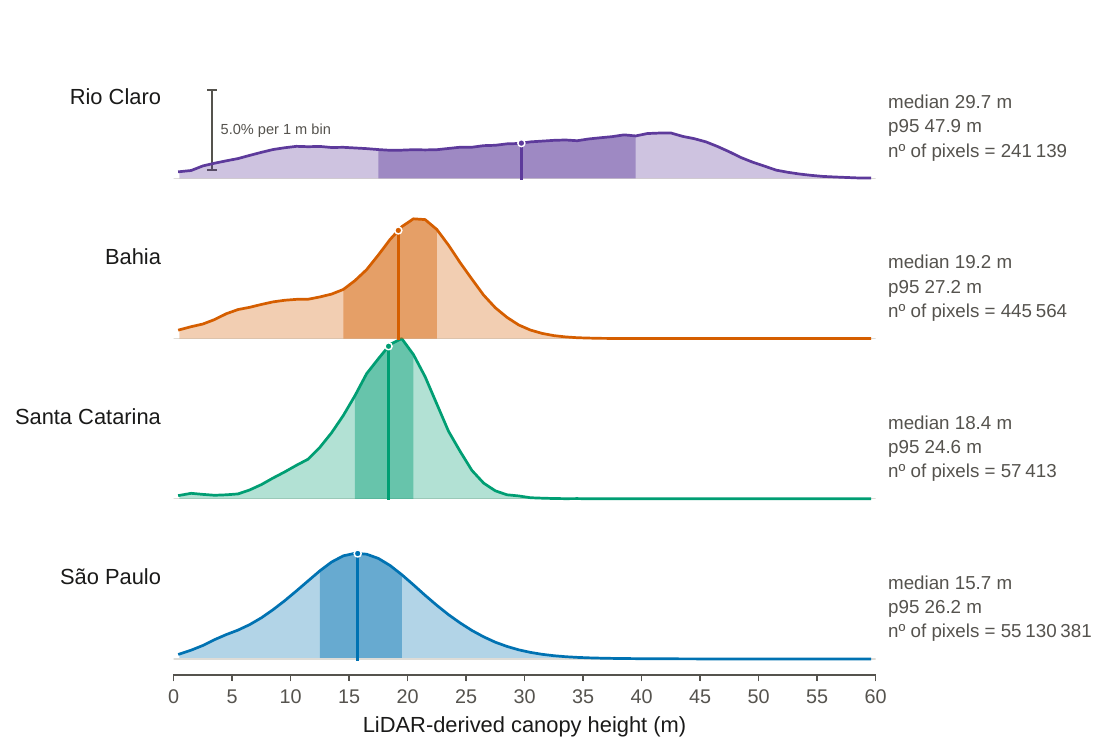}
\caption{Distribution of LiDAR canopy height in the S\~ao Paulo test patches and at the three external validation sites, Jaragu\'a do Sul in Santa Catarina, Serra do Conduru in Bahia and Rio Claro. Each band is a histogram in classes of 1~m over the pixels that enter the respective assessment, namely the pixels inside the MapBiomas forest mask with a height above zero in the S\~ao Paulo test patches, and all the pixels compared with the published product at the external sites, where no forest mask is applied. The interquartile range is shaded and the median is marked. The median, the 95th percentile and the number of pixels are reported on the right of each band. The vertical bar gives the share of pixels falling in a class of 1~m, the horizontal axis is truncated at 60~m and the bands are ordered by median height.}
\label{Fig:histCHM}
\end{figure}

\subsubsection{Comparison with the published product}

The assessment was made on the published product itself and not on a prediction made for the purpose. At each site we read the tiles of record 5 (``Data Records''), which hold the canopy height of 2024 before the forest mask is applied, and kept them on their own grid, without resampling. The LiDAR CHM was brought to that grid by area-weighted averaging, from the 10~m mosaics at Conduru and Jaragu\'a do Sul and from the 1~m CHM at Rio Claro. A cell entered the comparison when LiDAR covered at least 90\% of its area and its reference height was above zero, and at Rio Claro the cells were further restricted to the limits of the State Forest. The cells finally compared are those that also hold valid data in the global product of the next subsection, so that the two maps are assessed over the same pixels. No forest mask was applied, so that the plantations of Rio Claro are evaluated together with its native vegetation. The cells inside the forest mask of record 1 make up 94.5\% of the cells compared at Conduru, 98.2\% at Jaragu\'a do Sul and 61.8\% at Rio Claro, and inside that mask records 1 and 5 hold the same height. The product was compared with the LiDAR CHM cell by cell, with the metrics of ``Accuracy assessment''.

None of the three surveys shares a pixel with the training or the test patches, so the three sites measure the error of the product where the model had no reference data. The comparison is not contemporaneous, however. The product describes 2024, whereas the LiDAR was flown in 2015 at Conduru, in 2017 at Jaragu\'a do Sul and in 2026 at Rio Claro, and the pixels were not screened for change. Forest growth and disturbance during the nine and seven years that separate the first two surveys from the product are therefore part of the error measured there.

\subsection{Comparison with an existing canopy height product}
\label{sec:products}

We benchmarked our map against the global canopy height product of Meta and the World Resources Institute (WRI), in its second version, CHMv2 \citep{meta2026chm2}. The product maps canopy height worldwide at a native resolution of 1~m from very-high-resolution RGB imagery, with a self-supervised vision transformer backbone and a convolutional decoder trained on airborne LiDAR \citep{tolan2024very}. It is the finest global canopy height map available at the moment.

The comparison was made over two sets of pixels, the held-out test patches of S\~ao Paulo and the three external validation sites (``External validation sites''). In both, the tiles of the global product were read from its public repository and aggregated from their native 1~m grid to the 10~m grid of the comparison, that of the patches in S\~ao Paulo and that of record 5 at the external sites, by area-weighted averaging, the operator that also brought the LiDAR CHM to that grid. In the global product a height of zero is a valid value and a separate value marks the absence of data, so the zeros entered the average and only the 1~m pixels without data were left out of it. A 10~m cell was kept when valid data of the global product covered at least 90\% of its area. Only the cells valid in the LiDAR, in our map and in the global product were retained, so that the two maps are compared with the LiDAR CHM over exactly the same pixels, as a paired sample and not on separate footprints, with the metrics of ``Accuracy assessment''.

In the test patches, our map is the prediction of the four-branch model (E5) from the 2024 composites, and the reference, the forest mask and the restriction to reference heights above zero are those of ``Accuracy assessment''. The global product holds valid data in 99.9\% of the test pixels, which leaves 55{,}072{,}444 pixels in the comparison. According to the metadata of the tiles, the images behind the global product over the test patches were acquired between 2011 and 2019, and the most recent image of a patch dates from 2019 in 79\% of the patches, so the global product is at least three years older than the LiDAR campaign of 2023 and 2024. The test patches are also interspersed with the training patches (``Training and test split''), so this comparison evaluates our model inside the landscapes it was trained on, an advantage over a product built to work everywhere.

At the external sites, our map is the published product (record 5) and the comparison is independent of the training data of our model. The requirement of valid data in the global product removed 5.5\% of the cells with LiDAR at Conduru, where the global product has no data, and none at the other two sites. According to the metadata of the tiles, the images behind the global product were acquired between 2014 and 2018 at Conduru, in 2018 and 2019 at Jaragu\'a do Sul and in 2019 at Rio Claro. The global product is therefore within three years of the LiDAR at the first two sites and five years older than it at Rio Claro, whereas ours is nine and seven years more recent than the LiDAR at the first two sites and two years older than it at Rio Claro.

\subsection{Carbon accounting by natural domain}
\label{sec:domains_method}

In 2025, the Brazilian Institute of Geography and Statistics released an ecological regionalization of the national territory, \emph{Dom\'inios e Regi\~oes Naturais Terrestres dos Biomas do Brasil} \citep{ibge2025dominios}. It nests three levels, the biome, the natural domain, and the natural region, and its boundaries follow geology, geomorphology, soils, climate, and vegetation typology instead of administrative limits. Brazil is divided into 52 natural domains and 271 natural regions. The Atlantic Forest holds ten natural domains, six of them in the continental interior and four in the coastal system, together with 57 natural regions. We adopted this layer because it partitions the Atlantic Forest along the same physical axes that govern forest structure. The carbon map of ``Aboveground carbon density model'' was intersected with the ten domains on the 30~m grid.

\section*{Data Records}

The dataset is deposited at Zenodo under \url{https://doi.org/10.5281/zenodo.24233662} \citep{ferreira2026dataset}. All raster layers are GeoTIFF files in geographic coordinates (EPSG 4326, WGS 84), delivered as tiles that follow the grid of the Sentinel-2 mosaics used for prediction. Tiles are internally tiled and compressed with the DEFLATE algorithm and carry overviews, so that they can be read directly by GDAL-based software and by cloud-optimized GeoTIFF clients. Table~\ref{tab:records} lists the records, and Figure~\ref{Fig:mapsCHMAGCD} shows the canopy height and the carbon density layers over the whole Atlantic Forest region.

\begin{table}[H]
\centering
\caption{Records of the dataset. All rasters are GeoTIFF in EPSG 4326. Integer layers are scaled by 10 and use a no-data value of 65535, and floating-point layers use $-9999$.}
\label{tab:records}
\begin{adjustbox}{max width=\textwidth}
\begin{tabular}{llllll}
\toprule
Record & Content & Resolution & Format and type & Units & Extent \\
\midrule
1 & Canopy height, 2024 & 10~m & GeoTIFF, uint16, scale 0.1 & m & Native forest (MapBiomas) \\
2 & Aboveground carbon density, 2024 & 30~m & GeoTIFF, uint16, scale 0.1 & Mg~C~ha$^{-1}$ & Native forest (MapBiomas) \\
3 & Standard error of record 2 & 30~m & GeoTIFF, uint16, scale 0.1 & Mg~C~ha$^{-1}$ & Native forest (MapBiomas) \\
4 & Mean carbon density and its uncertainty & 1/120$^\circ$ ($\approx$1~km) & GeoTIFF, float32, 11 bands & Mg~C~ha$^{-1}$ (mean and standard-error terms), \% (relative errors), count & Cells with forest \\
5 & Canopy height, 2024, without forest mask & 10~m & GeoTIFF, uint16, scale 0.1 & m & Whole region, all land covers \\
6 & Aboveground carbon density, 2024, without forest mask & 30~m & GeoTIFF, uint16, scale 0.1 & Mg~C~ha$^{-1}$ & Whole region, all land covers \\
\bottomrule
\end{tabular}
\end{adjustbox}
\end{table}

\begin{figure}[!t]
\centering

\includegraphics[width=\textwidth]{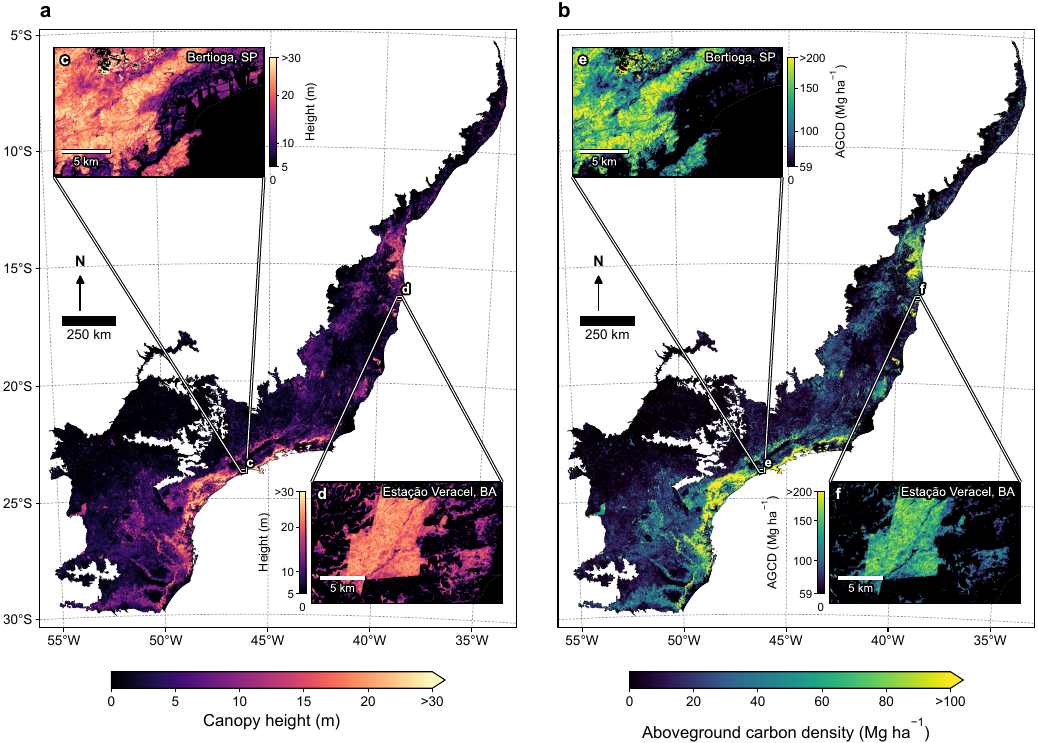}
\caption{Canopy height and aboveground carbon density of the Brazilian Atlantic Forest in 2024. (a) Canopy height (m) and (b) aboveground carbon density of record 2, both drawn over the Atlantic Forest region, filled in black, in an Albers equal-area projection. The insets show the same layers close to their native resolution at (c, e) the coast of Bertioga, in S\~ao Paulo, where the tall forest of the Serra do Mar escarpment gives way to the lower restinga and mangrove forests of the coastal plain, and (d, f) the Esta\c{c}\~ao Veracel, a private reserve of lowland rainforest in southern Bahia.}
\label{Fig:mapsCHMAGCD}
\end{figure}

\textbf{Record 1. Canopy height at 10~m.} One GeoTIFF per tile with the canopy height predicted by the four-branch model (configuration E5) from the annual 2024 composites, in metres, stored as unsigned 16-bit integers scaled by 10 (one unit equals 0.1~m) with a no-data value of 65535. The layer is masked to the MapBiomas forest classes, so that pixels outside native forest are no-data.

\textbf{Record 2. Aboveground carbon density at 30~m.} One GeoTIFF per tile with aboveground carbon density in Mg~C~ha$^{-1}$, obtained by aggregating the 10~m height to 30~m by spatial averaging and converting it with the power law of Equation~\ref{eq:agcd}. Values are stored as unsigned 16-bit integers scaled by 10 (one unit equals 0.1~Mg~C~ha$^{-1}$) with a no-data value of 65535, and the layer is masked to the MapBiomas forest classes, so that pixels outside native forest are no-data. This is the layer on which the carbon stocks of Table~\ref{tab:dominios} were computed.

\textbf{Record 3. Standard error of aboveground carbon density at 30~m.} One GeoTIFF per tile of record 2, on the same grid and over the same forest pixels, with the standard error of the carbon density of each pixel given by Equation~\ref{eq:var_pixel}, in Mg~C~ha$^{-1}$. Values are stored as unsigned 16-bit integers scaled by 10 with a no-data value of 65535, and the scale factor and the unit are written in the band metadata, so that GDAL-based software reads the physical value directly. The file metadata hold the coefficients of Equation~\ref{eq:agcd} used in the product ($a = 0.089$ and $b = 1.645$), the covariance matrix of the refit, and the variance function of Equation~\ref{eq:agcd_var}, so that each tile documents how it was computed. The relative standard error is 100 times this value divided by record 2. The standard error excludes the error of the predicted height (``Uncertainty of the carbon density map'') and is therefore a lower bound. Because the coefficient term is shared by all pixels, the pixel variances must not be summed as independent to obtain the uncertainty of a stock, for which record 4 should be used.

\textbf{Record 4. Mean carbon density and its uncertainty at 1~km.} One GeoTIFF on a grid of 1/120 of a degree with eleven bands, stored as 32-bit floating point with a no-data value of $-9999$. The bands hold the mean carbon density of the forest pixels of each cell, its standard error and relative standard error with the coefficient error shared by all pixels (Equation~\ref{eq:var_cell}) and, for comparison, with independent errors between pixels, the coefficient term of each of the two versions, the residual and within-cell terms, the standard deviation of the pixel values inside the cell and the number of forest pixels. The mean and all the uncertainty terms are in Mg~C~ha$^{-1}$, the terms being stored as standard errors, that is, as the square roots of the terms of Equation~\ref{eq:var_cell} and not as variances, and the relative errors are in per cent. Each band carries its name, unit, and description in its metadata. The correlated version is the one to use for stocks. A JSON file and a text file with the parameters of the uncertainty model accompany records 3 and 4.

\textbf{Record 5. Canopy height at 10~m without the forest mask.} The prediction of record 1 before the MapBiomas forest mask is applied, in the same format, so that every pixel holds a canopy height, including pastures, croplands, forest plantations, urban areas and water. It is delivered in 762 tiles that cover the whole Atlantic Forest region, 119 more than record 1, which keeps only the tiles with forest. The tiles extend beyond the limit of the region and over the sea and should be clipped to the area of interest. Inside the forest mask, record 5 is identical to record 1. The accuracy reported for the test patches under ``Technical Validation'' refers to forest pixels only. The external validation was run on this record without the forest mask and includes the eucalypt plantations of Rio Claro, but heights over pastures, croplands and urban areas remain essentially unvalidated.

\textbf{Record 6. Aboveground carbon density at 30~m without the forest mask.} Record 5 aggregated to 30~m and converted with Equation~\ref{eq:agcd} as in record 2, in the same format and on the same tiles as record 5. Inside the forest mask, it differs from record 2 only in the 30~m pixels that are partly forest, where record 2 averages only the forest pixels of the 10~m grid. Because Equation~\ref{eq:agcd} was calibrated on forest plots, outside native forest record 6 gives the carbon of a native stand of the mapped height and not an estimate of the carbon of plantations, crops or urban trees, and the standard errors of record 3 do not apply to it. It is distributed so that the carbon stock can be recomputed under another definition of forest.

\section{Technical Validation}
\label{sec:tv}

The validation proceeds from the calibration region outwards. We first report the accuracy of the five sensor configurations on the held-out test patches of S\~ao Paulo, the structure of the residuals by height class and the effect of aggregation to 30~m. We then compare the published canopy height with three independent airborne LiDAR surveys, side by side with the global canopy height product of Meta and WRI over the same pixels, repeat the comparison with that product over the test patches, document the fit and the residual structure of the height-to-carbon relationship, propagate its error into the carbon map from the pixel to the whole Atlantic Forest region, and finally test whether the carbon map reproduces the structural gradient of that region across an independent regionalization.

\subsection{Accuracy across sensor configurations}
\label{sec:results_accuracy}

Table~\ref{tab:accuracy_comparison_overall} reports RMSE, MAE, bias, $R^2$ and CCC for the five sensor configurations, computed on the held-out test patches described in ``Accuracy assessment'' and restricted to MapBiomas forest pixels with a reference LiDAR height above zero. E1 to E5 share the annual 2024 composite, the same patches, the same split and the same hyper-parameters, so the differences between them are attributable to the set of sensors alone.

\begin{table}[H]
\centering
\caption{Accuracy comparison of canopy height models (CHM) predicted from different sensor configurations, all with the annual 2024 composite, evaluated against airborne LiDAR reference height within the MapBiomas forest mask.}
\label{tab:accuracy_comparison_overall}
\begin{threeparttable}
\begin{adjustbox}{max width=\textwidth}
\begin{tabular}{llrrrrrr}
\toprule
ID & Sensors & RMSE (m) & MAE (m) & Bias (m) & Bias (\%) & $R^2$ & CCC \\
\midrule
E1 & S2 & 3.39 & 2.53 & +0.09 & +0.6 & 0.708 & 0.830 \\
E2 & S2 + S1 & 3.26 & 2.43 & \textbf{-0.04} & -0.3 & 0.732 & 0.852 \\
E3 & S2 + S1 + ALOS PALSAR & 3.14 & 2.33 & +0.09 & +0.5 & 0.749 & 0.858 \\
\addlinespace
E4 & S2 + S1 + ALOS DSM & 3.18 & 2.35 & +0.35 & +2.2 & 0.745 & 0.853 \\
E5 & S2 + S1 + ALOS PALSAR + ALOS DSM & \textbf{3.05} & \textbf{2.26} & +0.19 & +1.2 & \textbf{0.763} & 0.867 \\
\bottomrule
\end{tabular}
\end{adjustbox}
\end{threeparttable}
\end{table}

The optical baseline (E1) reached RMSE~=~3.39~m, MAE~=~2.53~m, $R^2$~=~0.708 and CCC~=~0.830. Adding the C-band branch (E2) produced the largest single gain of the series, lowering RMSE by 0.13~m and MAE by 0.10~m and raising $R^2$ to 0.732. The two branches added on top of that configuration behaved differently from one another. ALOS L-band backscatter (E3) lowered RMSE by a further 0.12~m and MAE by 0.10~m, almost as much as the C-band branch, and raised $R^2$ to 0.749. Elevation (E4) lowered RMSE and MAE by 0.08~m each and raised $R^2$ to 0.745, but it also shifted the pooled bias from $-$0.04 to +0.35~m, the largest of the series, so the DSM branch reduces the dispersion of the error less than the L-band branch and adds a systematic underestimation.

The four-branch model (E5) was the best configuration in RMSE, MAE, $R^2$ and CCC, with RMSE~=~3.05~m, MAE~=~2.26~m, $R^2$~=~0.763 and CCC~=~0.867, and its pooled bias of +0.19~m, or 1.2\% of the mean reference height, is larger than those of E1 to E3 and smaller than that of E4. Relative to the optical baseline, this amounts to a reduction of 9.8\% in RMSE and of 10.7\% in MAE. The four-branch model improved on E2 by 0.21~m in RMSE, close to the sum of the gains of E3 and E4 (0.20~m), so L-band backscatter and elevation carry largely complementary information. The improvement comes from combining a signal that responds to canopy structure with a description of the relief that modulates both the backscatter and the attainable height.

The comparison of Supplementary Table~\ref{tab:arch_comparison_E1_p10}, which trains the optical baseline with three architectures under identical conditions, supports the choice of the BPAT-UNet for the optical branch. The BPAT-UNet reached the lowest error, with RMSE~=~3.35~m, MAE~=~2.50~m, $R^2$~=~0.714 and CCC~=~0.835, against 3.38~m, 2.52~m, 0.711 and 0.835 for the MA-Net and 3.47~m, 2.60~m, 0.694 and 0.825 for the U-Net, although the U-Net had the smallest pooled bias, $-$0.06~m against +0.17~m. Its advantage over the MA-Net is small, 0.03~m in RMSE, but it was obtained without pre-training and with a third fewer parameters (43.6 against 64.9~M), whereas the U-Net, the lightest of the three (36.4~M), was the least accurate. The three networks share the height-dependent bias described below, overestimating canopy below 10~m by 1.6 to 1.9~m and underestimating canopy above 40~m by 17.4 to 17.9~m, so this compression of the upper tail does not depend on the architecture. As these networks were trained anew with a shorter patience, the BPAT-UNet row does not reproduce exactly the E1 row of Table~\ref{tab:accuracy_comparison_overall} (3.35 against 3.39~m in RMSE).

We stratified the residuals of the four-branch model (E5) by reference LiDAR height class (Fig.~\ref{Fig:biasByHeightAnnual}). The near-zero pooled bias of this model proves to be the residue of two large and opposite errors. Short canopy is systematically overestimated by 1.25~m in the 0 to 10~m class, which amounts to 18.7\% of the class mean height, and by 0.25~m (1.6\%) in the modal 10 to 20~m class, which alone holds 58.1\% of the evaluated pixels. Underestimation then grows monotonically above 20~m, reaching +2.0~m (8.6\%) at 20 to 30~m, +6.47~m (19.9\%) at 30 to 40~m, and +16.38~m (36.3\%) above 40~m.

\begin{figure}[H]
\centering
\includegraphics[width=0.85\textwidth]{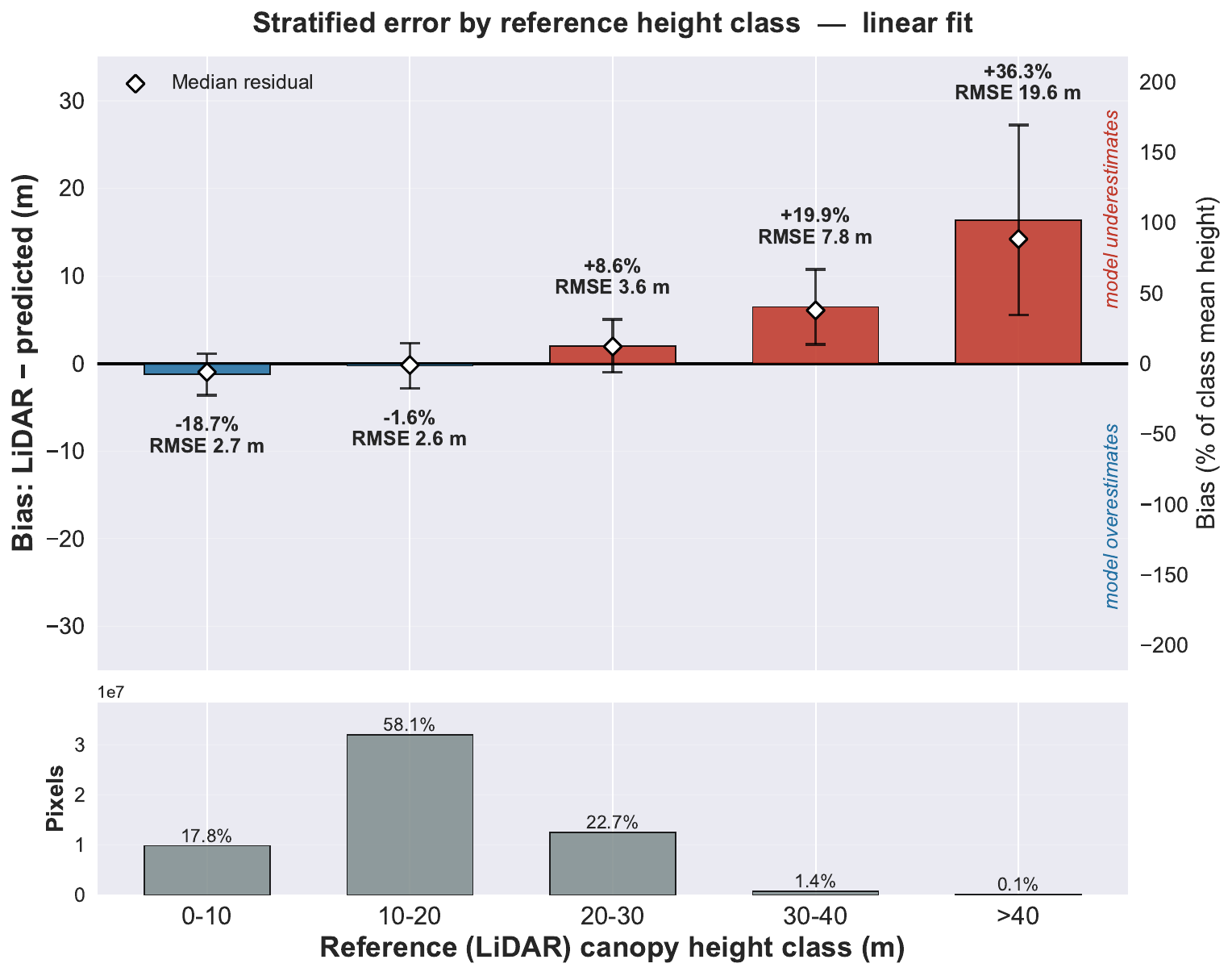}
\caption{Systematic error of the four-branch model (E5) by reference canopy height derived from LiDAR. Bias is the mean of LiDAR minus predicted height, so bars above zero, in red, denote underestimation and bars below zero, in blue, denote overestimation. The right-hand axis gives the same bias as a percentage of the class mean height, the label above or below each bar repeats that percentage together with the class RMSE, diamonds mark the median residual and error bars span $\pm$1~SD of the per-pixel residual within the class. The lower panel gives the number of test pixels in each class and its share of the total.}
\label{Fig:biasByHeightAnnual}
\end{figure}

Part of the pixel-level error is spatial noise rather than a systematic offset, and it disappears when the two maps are read at a coarser resolution. Aggregating both the prediction and the reference to a 30~m grid lowers RMSE from 3.05 to 2.58~m, and MAE from 2.26 to 1.90~m. The agreement statistics improve accordingly, with $R^2$ rising from 0.763 to 0.810 and CCC from 0.867 to 0.896. The bias, on the other hand, is practically untouched, +0.19~m at 10~m against +0.18~m at 30~m, as it must be, since averaging removes dispersion and not a systematic offset (Fig.~\ref{Fig:hexbin10mVs30m}). About 16\% of the error at the native resolution is therefore spatially uncorrelated and cancels within a 30~m pixel.

\begin{figure}[H]
\centering
\includegraphics[width=\textwidth]{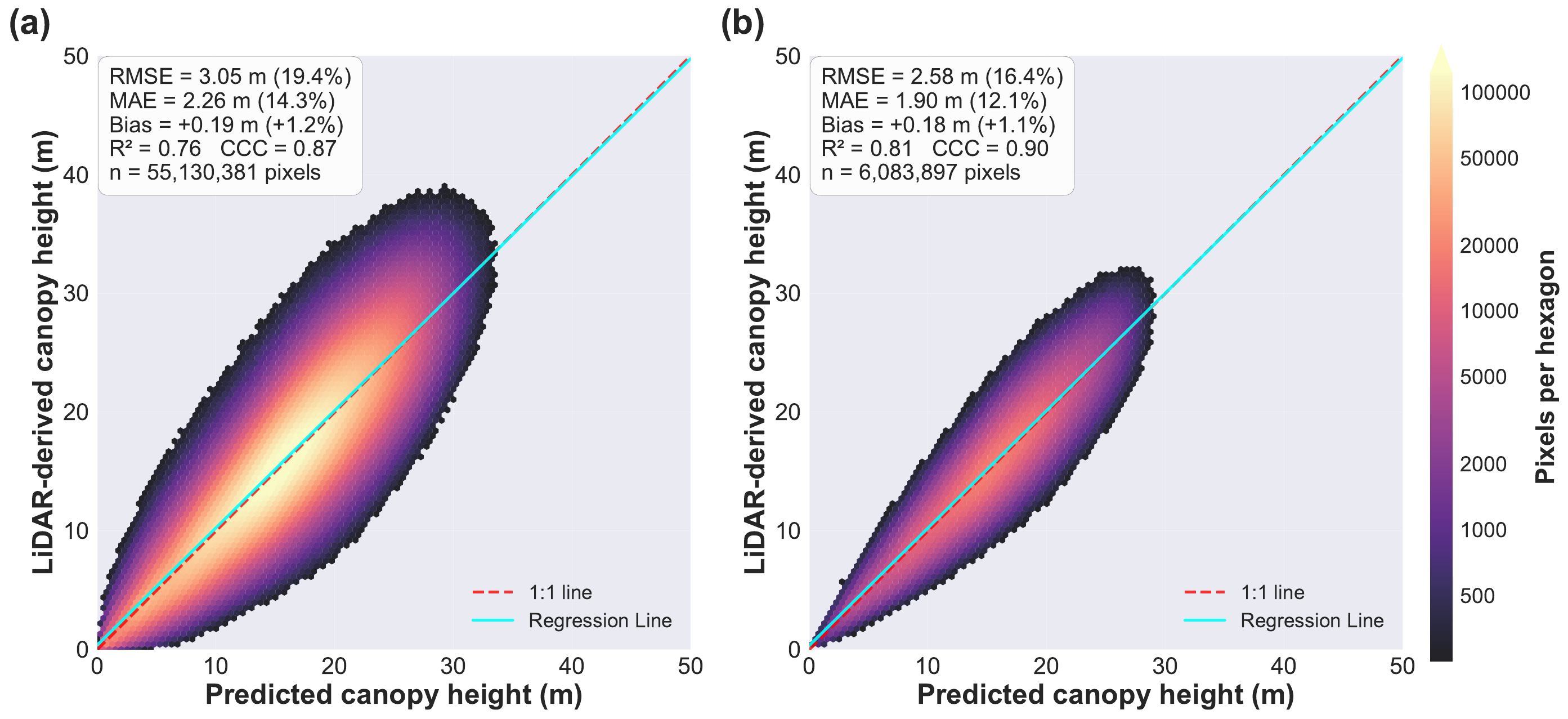}
\caption{Prediction versus LiDAR canopy height for the four-branch model (E5) at (a) the native 10~m resolution and (b) 30~m, after aggregating both the prediction and the reference by spatial averaging. The dashed line is the 1 to 1 line and the solid line is the ordinary least-squares fit, with bias defined as LiDAR~$-$~predicted.
Hexagon colour is the number of pixels per hexagon on a logarithmic scale shared by both panels and truncated at the 99th percentile (pointed end of the colour bar). Aggregation lowers RMSE and MAE and leaves the bias unchanged, which indicates that the error it removes is
spatially uncorrelated rather than a systematic offset. The values are given in the text.}
\label{Fig:hexbin10mVs30m}
\end{figure}

\subsection{External validation and comparison with the global product of Meta and WRI}
\label{sec:results_external}

The canopy height of record 5 was compared with the LiDAR CHM at the three sites of ``External validation sites'', and the global product of Meta and WRI was compared with the same LiDAR over the same pixels (``Comparison with an existing canopy height product''). Table~\ref{tab:external_overall} reports the overall metrics of the two maps and Figure~\ref{Fig:hexbinExternal} shows the reference height against the height of each map at each site.

\begin{table}[H]
\centering
\caption{External validation of the published canopy height (record 5, this study) and of the global product of Meta and WRI (CHMv2) at the three sites of ``External validation sites'', evaluated pixel by pixel on the 10~m grid of the product and over the same pixels for the two maps. The year of the LiDAR survey is given in brackets. Bias is the mean of the LiDAR reference minus the map, so positive values denote underestimation.}
\label{tab:external_overall}
\begin{adjustbox}{max width=\textwidth}
\begin{tabular}{llrrrrrrrrr}
\toprule
Site & Map & Pixels & Mean ref. (m) & RMSE (m) & RMSE (\%) & MAE (m) & Bias (m) & Bias (\%) & $R^2$ & CCC \\
\midrule
Serra do Conduru, BA (2015) & This study & 445{,}564 & 17.88 & 3.61 & 20.2 & 2.82 & +0.33 & +1.8 & 0.714 & 0.835 \\
 & Meta CHMv2 & 445{,}564 & 17.88 & 5.14 & 28.7 & 3.92 & +3.01 & +16.8 & 0.626 & 0.709 \\
\addlinespace
Jaragu\'a do Sul, SC (2017) & This study & 57{,}413 & 17.87 & 4.37 & 24.4 & 3.48 & $-$2.11 & $-$11.8 & 0.431 & 0.594 \\
 & Meta CHMv2 & 57{,}413 & 17.87 & 5.58 & 31.2 & 4.47 & $-$1.66 & $-$9.3 & 0.387 & 0.557 \\
\addlinespace
Rio Claro, SP (2026) & This study & 241{,}139 & 28.33 & 7.16 & 25.3 & 5.47 & +1.46 & +5.2 & 0.733 & 0.838 \\
 & Meta CHMv2 & 241{,}139 & 28.33 & 16.12 & 56.9 & 13.23 & +11.88 & +41.9 & 0.409 & 0.269 \\
\bottomrule
\end{tabular}
\end{adjustbox}
\end{table}

\begin{figure}[p]
\centering
\includegraphics[width=\textwidth,height=0.80\textheight,keepaspectratio]{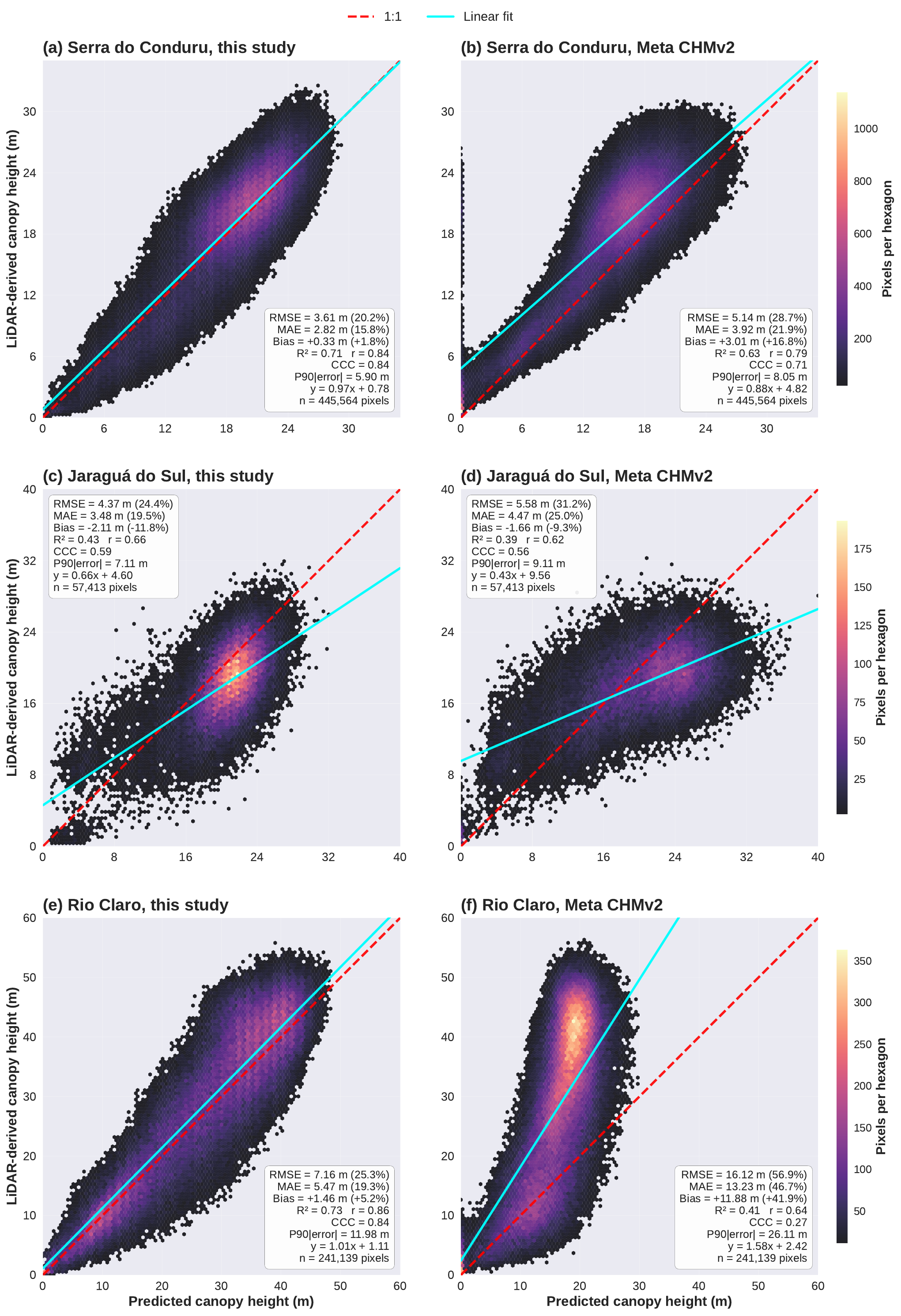}
\caption{External validation of the published canopy height (record 5) and comparison with the global product of Meta and WRI. LiDAR canopy height against the height of this study (a, c, e) and of Meta CHMv2 (b, d, f) at Serra do Conduru (a, b), Jaragu\'a do Sul (c, d) and Rio Claro (e, f), over the same pixels in the two panels of each row. Hexagon colour is the number of pixels per hexagon, on a scale shared by the two panels of a row, the dashed line is the 1 to 1 line and the solid line is the ordinary least-squares fit of reference on prediction. The axes differ between rows. Bias is LiDAR minus predicted height, r is the Pearson correlation coefficient and P90 is the 90th percentile of the absolute error.}
\label{Fig:hexbinExternal}
\end{figure}

At Serra do Conduru, the comparison covers 445{,}564 pixels with a mean reference height of 17.88~m. The published product has an RMSE of 3.61~m, which is 20.2\% of that mean, an MAE of 2.82~m (15.8\%) and a pooled bias of +0.33~m (1.8\%), with a median bias of +0.31~m. Half of the pixels fall within 2.32~m of the reference and 90\% within 5.90~m. $R^2$ is 0.714, and CCC is 0.835. At Jaragu\'a do Sul the comparison covers 57{,}413 pixels with a mean reference height of 17.87~m. RMSE is 4.37~m (24.4\%), MAE is 3.48~m (19.5\%), and the pooled bias is $-$2.11~m ($-$11.8\%), with a median bias of $-$2.21~m, so the product overestimates the canopy of this site. Half of the pixels fall within 2.96~m of the reference and 90\% within 7.11~m. $R^2$ is 0.431, and CCC is 0.594, the weakest agreement of the three sites, over the narrowest range of reference heights. At Rio Claro the comparison covers 241{,}139 pixels with a mean reference height of 28.33~m, about 10~m above that of the other two sites. RMSE is 7.16~m (25.3\%), MAE is 5.47~m (19.3\%), and the pooled bias is +1.46~m (5.2\%), with a median bias of +1.24~m. Half of the pixels fall within 4.28~m of the reference and 90\% within 11.98~m. $R^2$ is 0.733, and CCC is 0.838. The absolute error is largest at Rio Claro because its canopy is the tallest. Relative to the mean reference height the three sites are close to one another, with an RMSE of 20 to 25\% and an MAE of 16 to 19\%.

Records 1 and 5 hold the same height inside the forest mask, so the comparison restricted to the cells inside that mask evaluates record 1. Over those cells, which are 94.5\%, 98.2\% and 61.8\% of the cells compared at the three sites, the MAE of the product is 2.85, 3.47 and 5.46~m and its pooled bias is +0.40, $-$2.20 and +1.12~m. The MAE stays within 0.03~m of the value obtained over all the cells, so the forest mask does not change the assessment.

Over the same pixels, the global product of Meta and WRI has a larger error than the published product at the three sites. Its MAE is 3.92~m at Conduru, 4.47~m at Jaragu\'a do Sul and 13.23~m at Rio Claro, which is 1.4, 1.3 and 2.4 times ours, and its RMSE is 5.14, 5.58 and 16.12~m. At Conduru most of the difference is systematic. The global product underestimates the canopy by 3.01~m (16.8\%), against 0.33~m for ours, and in 3{,}950 pixels (0.9\%) it maps a height below 1~m under a LiDAR canopy taller than 10~m, which produces the vertical alignment of points at zero in panel (b) of Figure~\ref{Fig:hexbinExternal}. At Jaragu\'a do Sul the two maps overestimate the canopy by similar amounts, 1.66~m for the global product and 2.11~m for ours, and the larger error of the global product comes from its dispersion, with a standard deviation of the residuals of 5.33~m against 3.83~m. At Rio Claro the global product reaches a mean height of 16.4~m where the LiDAR measures 28.3~m, an underestimation of 11.88~m (41.9\%). Its CCC falls to 0.269 and the 90th percentile of its absolute error is 26.11~m, against 11.98~m for ours. The LiDAR places 49.4\% of the pixels of Rio Claro above 30~m, the published product 46.3\% and the global product 0.5\%. Over the pixels that the LiDAR places above 30~m, the height of the global product has a median of 19.1~m and stays between 13.9 and 25.5~m in nine pixels out of ten, a pattern of saturation. The images behind the global product at Rio Claro date from 2019, seven years before the LiDAR survey, so growth in the interval may account for part of this difference, although the stands are old \citep{fflorestal2024feena} and the deficit is concentrated in the tallest canopy.

The difference between the two maps at Rio Claro is not an effect of registration. When each map is shifted by up to two pixels in each direction against the LiDAR, the published product correlates best without any shift at the three sites, and so does the global product at Conduru and Jaragu\'a do Sul. At Rio Claro the global product correlates best when displaced by one pixel, which raises its correlation coefficient from 0.640 to 0.721 but lowers its RMSE only to 15.75~m and leaves its bias at +11.90~m, so its error there is one of height and not of position.

Table~\ref{tab:bias_by_height_all} stratifies the residuals of the two maps by reference height class at the three sites and, for comparison, in the S\~ao Paulo test patches, where the map of this study is the prediction of the four-branch model. At the external sites the published product repeats the pattern of the test patches, in which short canopy is overestimated and tall canopy is underestimated. Its bias goes from $-$2.13~m ($-$34.5\%) between 0 and 10~m to +6.52~m (20.5\%) between 30 and 40~m at Conduru, from $-$3.50~m ($-$49.9\%) between 0 and 10~m to +0.10~m (0.4\%) between 20 and 30~m at Jaragu\'a do Sul, and from $-$2.83~m ($-$45.5\%) between 0 and 10~m to +7.41~m (16.4\%) above 40~m at Rio Claro. Above 20~m the bias at Conduru matches that of the test patches, with +1.93 against +2.00~m between 20 and 30~m and +6.52 against +6.47~m between 30 and 40~m. At Rio Claro the upper tail is less compressed than in the test patches, with +2.14 against +6.47~m between 30 and 40~m and +7.41 against +16.38~m above 40~m. Below 20~m the product overestimates the canopy more at the three sites than in the test patches. At Conduru and Jaragu\'a do Sul part of this overestimation may be real, since the short stands measured by the LiDAR had nine and seven years to grow before the year of the product, and the comparison cannot separate that growth from the error of the map.

The underestimation of tall canopy is stronger in the global product. Its bias grows from +0.60~m between 0 and 10~m to +8.55~m (26.9\%) between 30 and 40~m at Conduru, and at Rio Claro it reaches +8.54~m (33.9\%) between 20 and 30~m, +16.34~m (46.5\%) between 30 and 40~m and +25.20~m (55.9\%) above 40~m. Between 0 and 10~m the global product is closer to the LiDAR than ours at Conduru and Jaragu\'a do Sul, with a bias of +0.60 and $-$0.24~m, and at Rio Claro it overestimates the short canopy about as much as ours does, by 2.47 against 2.83~m. The residuals of the global product in the test patches are examined in the next subsection.

\begin{table}[H]
\centering
\caption{Systematic error by reference canopy height class in the S\~ao Paulo test patches and at the three external sites, for this study and for the global product of Meta and WRI (CHMv2). This study is the four-branch model (E5) in the test patches and the published product (record 5) at the external sites. Bias is the mean of the LiDAR reference minus the map, so positive values denote underestimation and negative values denote overestimation, and the value in brackets is the same bias as a percentage of the class mean height. The middle panel gives the share of the evaluated pixels that falls in each class and the last row gives the number of evaluated pixels. In the test patches the global product is evaluated over the 99.9\% of the pixels in which it holds valid data, and over that subset the bias of the model stays within 0.01~m of the values shown. Classes holding fewer than one hundred pixels are marked n.a.}
\label{tab:bias_by_height_all}
\begin{adjustbox}{max width=\textwidth}
\begin{tabular}{lcccccccc}
\toprule
 & \multicolumn{2}{c}{S\~ao Paulo (test)} & \multicolumn{2}{c}{Serra do Conduru, BA} & \multicolumn{2}{c}{Jaragu\'a do Sul, SC} & \multicolumn{2}{c}{Rio Claro, SP} \\
\cmidrule(lr){2-3}\cmidrule(lr){4-5}\cmidrule(lr){6-7}\cmidrule(lr){8-9}
Class (m) & This study & Meta CHMv2 & This study & Meta CHMv2 & This study & Meta CHMv2 & This study & Meta CHMv2 \\
\midrule
\multicolumn{9}{l}{\emph{Bias in m, and in \% of the class mean height}} \\
0 to 10   & $-$1.25 ($-$18.7) & $-$1.46 ($-$21.8) & $-$2.13 ($-$34.5) & +0.60 (+9.7) & $-$3.50 ($-$49.9) & $-$0.24 ($-$3.5) & $-$2.83 ($-$45.5) & $-$2.47 ($-$39.8) \\
10 to 20  & $-$0.25 ($-$1.6) & $-$1.19 ($-$7.8) & $-$0.64 ($-$4.0) & +1.73 (+10.7) & $-$3.23 ($-$19.7) & $-$2.46 ($-$15.1) & $-$2.22 ($-$14.9) & +1.32 (+8.9) \\
20 to 30  & +2.00 (+8.6) & +2.85 (+12.2) & +1.93 (+8.3) & +4.90 (+21.0) & +0.10 (+0.4) & $-$0.52 ($-$2.3) & $-$0.64 ($-$2.6) & +8.54 (+33.9) \\
30 to 40  & +6.47 (+19.9) & +8.58 (+26.4) & +6.52 (+20.5) & +8.55 (+26.9) & n.a. & n.a. & +2.14 (+6.1) & +16.34 (+46.5) \\
$>$40     & +16.38 (+36.3) & +19.36 (+42.9) & n.a. & n.a. & n.a. & n.a. & +7.41 (+16.4) & +25.20 (+55.9) \\
\addlinespace
\multicolumn{9}{l}{\emph{Share of the evaluated pixels, in \%}} \\
0 to 10   & \multicolumn{2}{c}{17.8} & \multicolumn{2}{c}{15.6} & \multicolumn{2}{c}{6.1} & \multicolumn{2}{c}{11.8} \\
10 to 20  & \multicolumn{2}{c}{58.1} & \multicolumn{2}{c}{40.0} & \multicolumn{2}{c}{60.2} & \multicolumn{2}{c}{19.1} \\
20 to 30  & \multicolumn{2}{c}{22.7} & \multicolumn{2}{c}{43.1} & \multicolumn{2}{c}{33.6} & \multicolumn{2}{c}{19.7} \\
30 to 40  & \multicolumn{2}{c}{1.4} & \multicolumn{2}{c}{1.3} & \multicolumn{2}{c}{0.2} & \multicolumn{2}{c}{24.9} \\
$>$40     & \multicolumn{2}{c}{0.1} & \multicolumn{2}{c}{0.002} & \multicolumn{2}{c}{0.01} & \multicolumn{2}{c}{24.5} \\
\addlinespace
\midrule
Evaluated pixels & 55{,}130{,}381 & 55{,}072{,}444 & \multicolumn{2}{c}{445{,}564} & \multicolumn{2}{c}{57{,}413} & \multicolumn{2}{c}{241{,}139} \\
\bottomrule
\end{tabular}
\end{adjustbox}
\end{table}

\subsection{Comparison with the global product over the test patches}
\label{sec:results_meta}

The comparison with the global product was repeated over the held-out test patches of S\~ao Paulo, with the prediction of the four-branch model (E5) in place of the published product (``Comparison with an existing canopy height product''). This sample is far larger than that of the external sites and is spread over the whole State. The global product holds valid data in 55{,}072{,}444 of the 55{,}130{,}381 test pixels (99.9\%), and the two maps were compared over these pixels (Fig.~\ref{Fig:hexbinMeta}). The four-branch model has RMSE~=~3.05~m (19.4\%), MAE~=~2.26~m (14.3\%), bias~=~+0.19~m (1.2\%), $R^2$~=~0.763 and CCC~=~0.866, within 0.001 of the values of Table~\ref{tab:accuracy_comparison_overall}, which refer to all the test pixels. The global product has RMSE~=~5.39~m (34.1\%), MAE~=~4.14~m (26.3\%), bias~=~$-$0.17~m ($-$1.1\%), $R^2$~=~0.426 and CCC~=~0.651. The 90th percentile of its absolute error is 8.61~m, against 4.79~m for our model. The error of the global product is thus 1.8 times ours, both in MAE and in RMSE.

\begin{figure}[!htb]
\centering
\includegraphics[width=\textwidth]{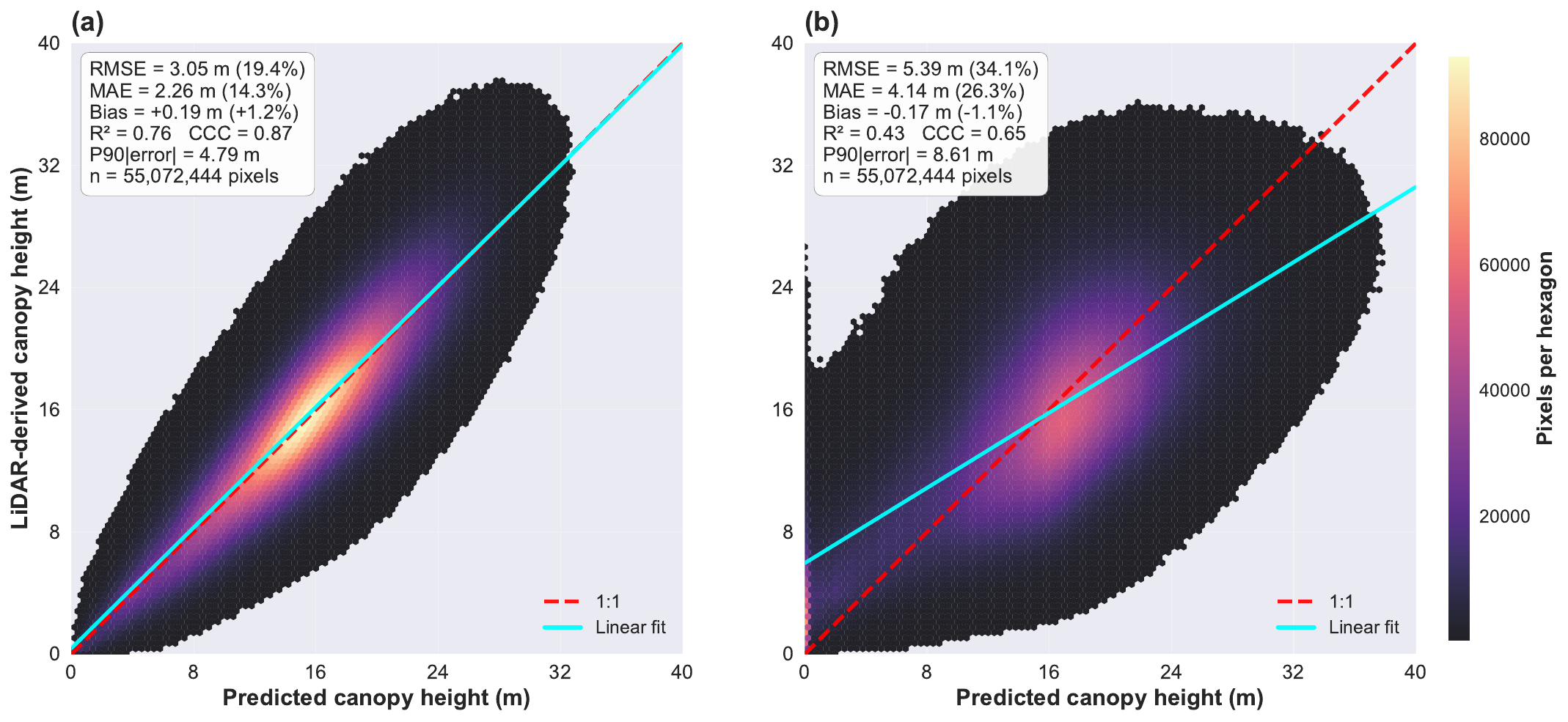}
\caption{Comparison with the global canopy height product of Meta and WRI (CHMv2) over the test patches of S\~ao Paulo. LiDAR canopy height against (a) the prediction of the four-branch model (E5) and (b) the global product, over the same 55{,}072{,}444 pixels, inside the MapBiomas forest mask and with a reference height above zero. Hexagon colour is the number of pixels per hexagon, on a linear scale shared by the two panels, and hexagons with fewer than 250 pixels are not drawn. The dashed line is the 1 to 1 line and the solid line is the ordinary least-squares fit of reference on prediction. Bias is LiDAR minus predicted height and P90 is the 90th percentile of the absolute error.}
\label{Fig:hexbinMeta}
\end{figure}

Unlike at Conduru and Rio Claro, the larger error of the global product in the test patches is not a systematic offset. Its pooled bias is close to zero, and the standard deviation of its residuals is 5.38~m, against 3.05~m for our model. By reference height class (Table~\ref{tab:bias_by_height_all}), the global product overestimates the canopy below 20~m, by 1.46~m (21.8\%) between 0 and 10~m and by 1.19~m (7.8\%) between 10 and 20~m, and underestimates it above, by 2.85~m (12.2\%) between 20 and 30~m, by 8.58~m (26.4\%) between 30 and 40~m and by 19.36~m (42.9\%) above 40~m. This is the pattern of our model, with a larger bias in every class and with more dispersion within the classes. The RMSE of the three classes below 30~m, which hold 98.6\% of the pixels, is 4.77 to 6.19~m for the global product and 2.61 to 3.63~m for ours. The global product does not compress the range of heights, since it places 1.56\% of the pixels above 30~m, close to the 1.44\% of the LiDAR, whereas our model places 0.35\%. Its larger error in the test patches therefore comes from the dispersion of its estimates around the reference. In 140{,}698 pixels (0.3\%) the global product maps a height below 1~m under a LiDAR canopy taller than 10~m, which produces the vertical alignment of points at the left of panel (b) of Figure~\ref{Fig:hexbinMeta}.

Two circumstances favour our model in this comparison. The test patches are interspersed with the training patches (``Training and test split''), so the model is evaluated inside the landscapes it was trained on, whereas the global product was built to work everywhere. The images behind the global product over the test patches are also at least three years older than the LiDAR campaign, whereas our composites date from 2024 (``Comparison with an existing canopy height product''). The comparison at the three external sites, which share no pixel with the training or the test patches and where the dates favour the global product at Conduru and Jaragu\'a do Sul, is therefore the more demanding test of our map. The two comparisons agree. The MAE of the global product is 1.8 times ours in the test patches and 1.3 to 2.4 times ours at the external sites.

\subsection{Carbon density as a function of canopy height}
\label{sec:results_agcd}

The 124 inventory plots span top canopy heights from 0.7 to 24.0~m and carbon densities from 0.06 to 19.3~kg~m$^{-2}$, with means of 11.7~m and 5.70~kg~m$^{-2}$. Equation~\ref{eq:agcd} was adopted with $a = 0.089$ and $b = 1.645$, and over these plots it explains 76\% of the variance of plot carbon density, with an RMSE of 1.90~kg~m$^{-2}$, an MAE of 1.43~kg~m$^{-2}$ and a bias of $+$0.11~kg~m$^{-2}$ (Fig.~\ref{Fig:agcdModel}).

\begin{figure}[H]
\centering

\includegraphics[width=\textwidth]{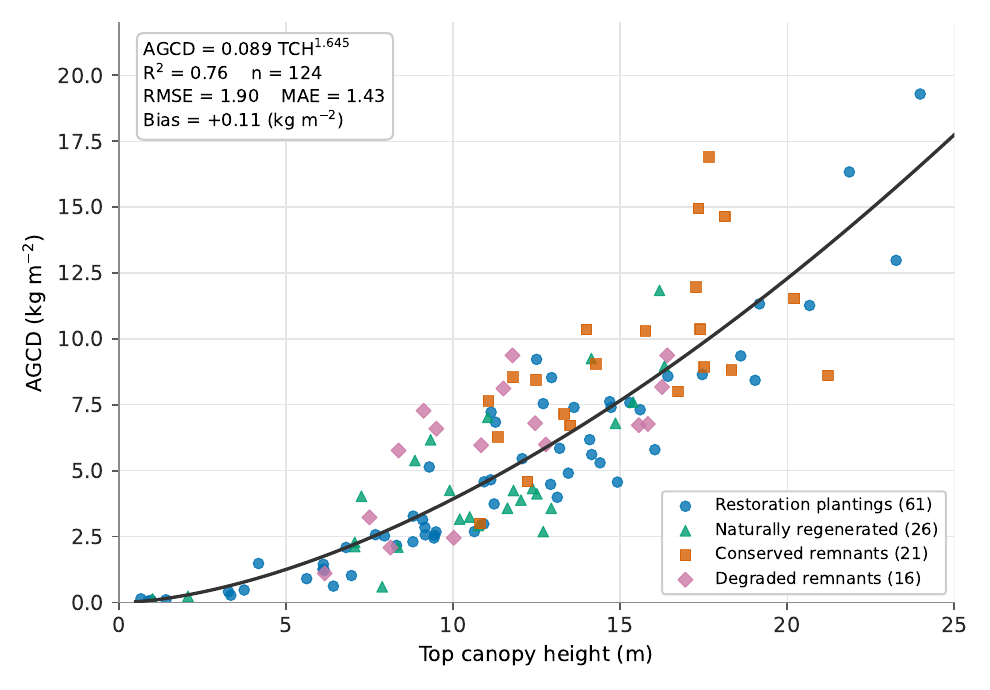}
\caption{Aboveground carbon density (AGCD) as a function of top canopy height (TCH) for the 124 inventory plots, with the plots coloured by category. The solid line is Equation~\ref{eq:agcd} with $a = 0.089$ and $b = 1.645$, and the box gives these coefficients together with the goodness-of-fit statistics evaluated on the plots, all in kg~m$^{-2}$, where 1~kg~m$^{-2}$ equals 10~Mg~C~ha$^{-1}$. Bias is the mean of observed minus predicted carbon density, so a positive value means that the curve lies below the plots. The number of plots in each category is given in the legend.}
\label{Fig:agcdModel}
\end{figure}

\subsection{Uncertainty of carbon density from the pixel to the whole region}
\label{sec:results_uncertainty}

The fit of Equation~\ref{eq:agcd_var} gave $a = 0.0898$ (SE 0.0186) and $b = 1.650$ (SE 0.080), with $R^2 = 0.758$, an RMSE of 1.90~kg~m$^{-2}$ and a residual standard deviation of $\sigma(\mathrm{TCH}) = 0.094\,\mathrm{TCH}^{\,1.197}$~kg~m$^{-2}$. The residual term accounts for 97.3\% of the variance of the pixels and the coefficients for the remaining 2.7\% (Equation~\ref{eq:var_pixel}). This high proportion of residual variance is typical in AGCD modelling. It indicates that while the mean height-to-biomass relationship is firmly established (yielding low coefficient uncertainty), individual pixels of the same height still exhibit natural variations in biomass due to unobserved factors like wood density and species composition.

The standard error of a pixel grows with canopy height, from 6.5~Mg~C~ha$^{-1}$ at 5~m to 14.9~Mg~C~ha$^{-1}$ at 10~m and 34.4~Mg~C~ha$^{-1}$ at 20~m, while the relative standard error falls from 52\% to 38\% and 28\% (Fig.~\ref{Fig:agcdUncertainty}b), because the residual grows with an exponent of 1.197 and the curve with one of 1.645. Over all forest pixels, the median standard error is 21.1~Mg~C~ha$^{-1}$, with 5th and 95th percentiles of 4.6 and 39.6~Mg~C~ha$^{-1}$, and the median relative standard error is 33.1\%, with percentiles of 26.8 and 60.1\%. Only 2.66\% of the forest pixels have a mapped height above the tallest plot, 24.0~m, where the standard error extrapolates the variance function.

Because the standard error depends only on height, its map mirrors the canopy height of Fig.~\ref{Fig:mapsCHMAGCD} (Fig.~\ref{Fig:agcdUncertainty}a). The highest values, above 30~Mg~C~ha$^{-1}$, occur in the tall continuous forests of the Serra do Mar in S\~ao Paulo and Paran\'a, of the coast of Santa Catarina and of southern Bahia, and the lowest in the shorter seasonal forests of western S\~ao Paulo and Mato Grosso do Sul, where the relative error is in turn the highest. For the whole Atlantic Forest region, the standard error of the mean density is 2.4~Mg~C~ha$^{-1}$, or 3.5\%, almost all of it from the coefficients, which corresponds to about 68~Tg~C on the stock of Table~\ref{tab:dominios}.

\begin{figure}[!t]
\centering
\includegraphics[width=110mm]{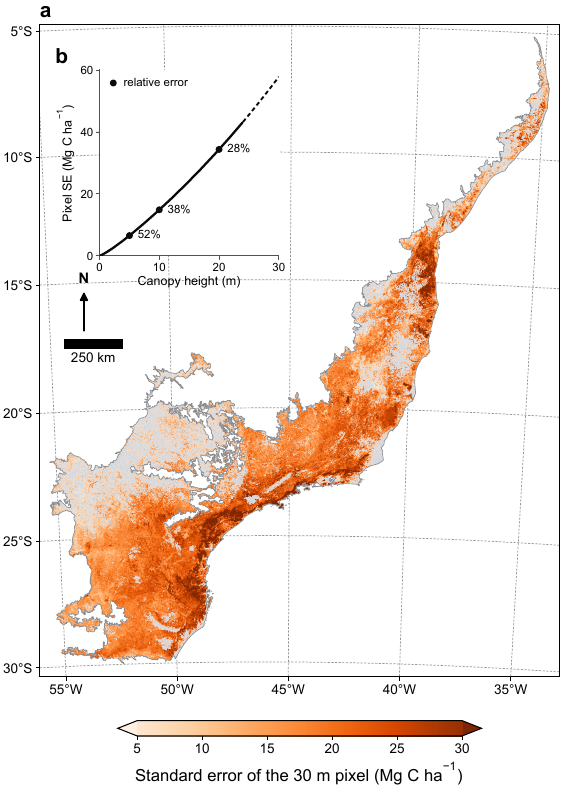}
\caption{Standard error of the aboveground carbon density (AGCD) map. (a) Standard error of the 30~m pixels (Equation~\ref{eq:var_pixel}, record 3), averaged over the forest pixels of display cells of about 1.2~km. Grey marks the parts of the region where forest covers less than 3\% of the cell. (b) Standard error of a pixel as a function of canopy height, solid within the height range of the 124 plots and dashed beyond the tallest plot (24.0~m), with dots giving the relative standard error at 5, 10 and 20~m. The error of the predicted height is not included.}
\label{Fig:agcdUncertainty}
\end{figure}

\subsection{Carbon stock across the natural domains}
\label{sec:results_dominios}

We analysed the carbon map in ten natural domains of the Atlantic Forest defined by the Brazilian Institute of Geography and Statistics \citep{ibge2025dominios} (``Carbon accounting by natural domain''). Table~\ref{tab:dominios} and Figure~\ref{Fig:dominios} summarize the results.

The forest mapped inside the ten domains covers 28.3~Mha and holds 1{,}944~Tg~C, for an overall mean of 68.8~Mg~C~ha$^{-1}$. A further 0.32~Mha of mapped forest, 1.1\% of the total, lies inside the boundary of the Atlantic Forest region but outside the domain layer and is left out of the accounting.

The stock is concentrated. Three domains, the Dense Forest of the Eastern Southeast, the Subtropical Mixed Forest, and the Dense Forest of the East and Eastern Northeast, hold 1{,}253~Tg~C, which is 64.5\% of the total stock, on 55.7\% of the total forest area. The Dense Forest of the Eastern Southeast is the densest and by far the best conserved of the ten, with 55.6\% of its area still under forest and a mean of 98.8~Mg~C~ha$^{-1}$, so 5.04~Mha carry 498~Tg~C. The Subtropical Mixed Forest of the Araucaria region carries the largest forest area of the ten domains, 6.82~Mha, at a mean of 67.0~Mg~C~ha$^{-1}$, and reaches 457~Tg~C with a third more area and a third less density. At the other end, the Semideciduous Forest of the Southern West retains only 10.0\% of its area under forest at a mean of 39.6~Mg~C~ha$^{-1}$, so it contributes 4.5\% of the stock from 7.9\% of the forest area. The four coastal domains together hold 0.80~Mha and 46~Tg~C, 2.4\% of the total stock, which reflects how narrow they are rather than how poor, since the Eastern Southeast Coast reaches a mean of 71.1~Mg~C~ha$^{-1}$, above six of the continental domains.

The shape of the distributions separates the domains as clearly as their means (Fig.~\ref{Fig:dominios}b). In the two dense forest domains the distribution is broad and close to symmetric, with an interquartile range of 66.2 to 129.7~Mg~C~ha$^{-1}$ in the Eastern Southeast and of 35.6 to 111.6 in the East and Eastern Northeast, and with 5\% of the area above 177 and 164~Mg~C~ha$^{-1}$ respectively. In the drier and more fragmented domains, the distribution collapses towards the origin and acquires a long right tail, so that the Eastern Northeast Coast has a median of 20.8~Mg~C~ha$^{-1}$ and the Semideciduous Forest of the Southern West a median of 36.7, while their 95th percentiles still reach 71 and 81. The gradient of the medians, from 20.8 to 98.5~Mg~C~ha$^{-1}$, covers almost a factor of five across a single biogeographic region, and it follows the ordering of forest cover more closely than the ordering of domain area, which is what the fragmentation history of the Atlantic Forest would predict.

\begin{table}[H]
\centering
\caption{Aboveground carbon in the remaining forest of the ten natural domains of the Atlantic Forest \citep{ibge2025dominios}, from the 30~m AGCD map inside the MapBiomas forest mask. The number in the first column is the one used in Figure~\ref{Fig:dominios} and orders the domains by median AGCD. Cover is the share of the domain area that the forest mask retains. All areas, stocks and quantiles weight each pixel by its true ground area. Cover, shares and totals were computed from unrounded values, so they can differ in the last digit from the ratios and sums of the rounded entries.}
\label{tab:dominios}
\begin{threeparttable}
\begin{adjustbox}{max width=\textwidth}
\begin{tabular}{lllrrrrrrrr}
\toprule
\# & Natural domain & System & Domain & Forest & Cover & Mean & Median & P95 & Stock & Share \\
 & & & (Mha) & (Mha) & (\%) & \multicolumn{3}{c}{(Mg C ha$^{-1}$)} & (Tg C) & (\%) \\
\midrule
1  & Dense Forest of the Eastern Southeast         & Interior & 9.06  & 5.04 & 55.6 & 98.8 & 98.5 & 176.8 & 498.3 & 25.6 \\
2  & Dense Forest of the East and Eastern Northeast & Interior & 15.83 & 3.88 & 24.5 & 76.7 & 72.0 & 164.0 & 297.6 & 15.3 \\
3  & Subtropical Mixed Forest, Araucaria Forests   & Interior & 19.46 & 6.82 & 35.1 & 67.0 & 67.4 & 123.1 & 457.2 & 23.5 \\
4  & Eastern Southeast Coast                       & Coastal  & 0.91  & 0.47 & 51.4 & 71.1 & 64.9 & 156.0 & 33.2  & 1.7  \\
5  & Transition Areas between Domains              & Interior & 21.20 & 4.99 & 23.6 & 58.7 & 55.6 & 122.8 & 293.2 & 15.1 \\
6  & Semideciduous Forest of the Central Southeast & Interior & 19.32 & 4.49 & 23.2 & 58.7 & 54.6 & 127.4 & 263.1 & 13.5 \\
7  & Northern South Coast                          & Coastal  & 0.25  & 0.01 & 5.5  & 53.3 & 45.6 & 124.6 & 0.7   & 0.0  \\
8  & Semideciduous Forest of the Southern West     & Interior & 22.26 & 2.23 & 10.0 & 39.6 & 36.7 & 81.3  & 88.2  & 4.5  \\
9  & Eastern Coast                                 & Coastal  & 1.10  & 0.22 & 20.0 & 44.1 & 34.6 & 114.6 & 9.7   & 0.5  \\
10 & Eastern Northeast Coast                       & Coastal  & 0.53  & 0.10 & 19.3 & 26.8 & 20.8 & 71.3  & 2.8   & 0.1  \\
\midrule
   & Atlantic Forest                               &          & 109.93 & 28.26 & 25.7 & 68.8 & 64.2 & 148.2 & 1944.1 & 100 \\
\bottomrule
\end{tabular}
\end{adjustbox}
\end{threeparttable}
\end{table}

\begin{figure}[!t]
\centering
\includegraphics[width=\textwidth]{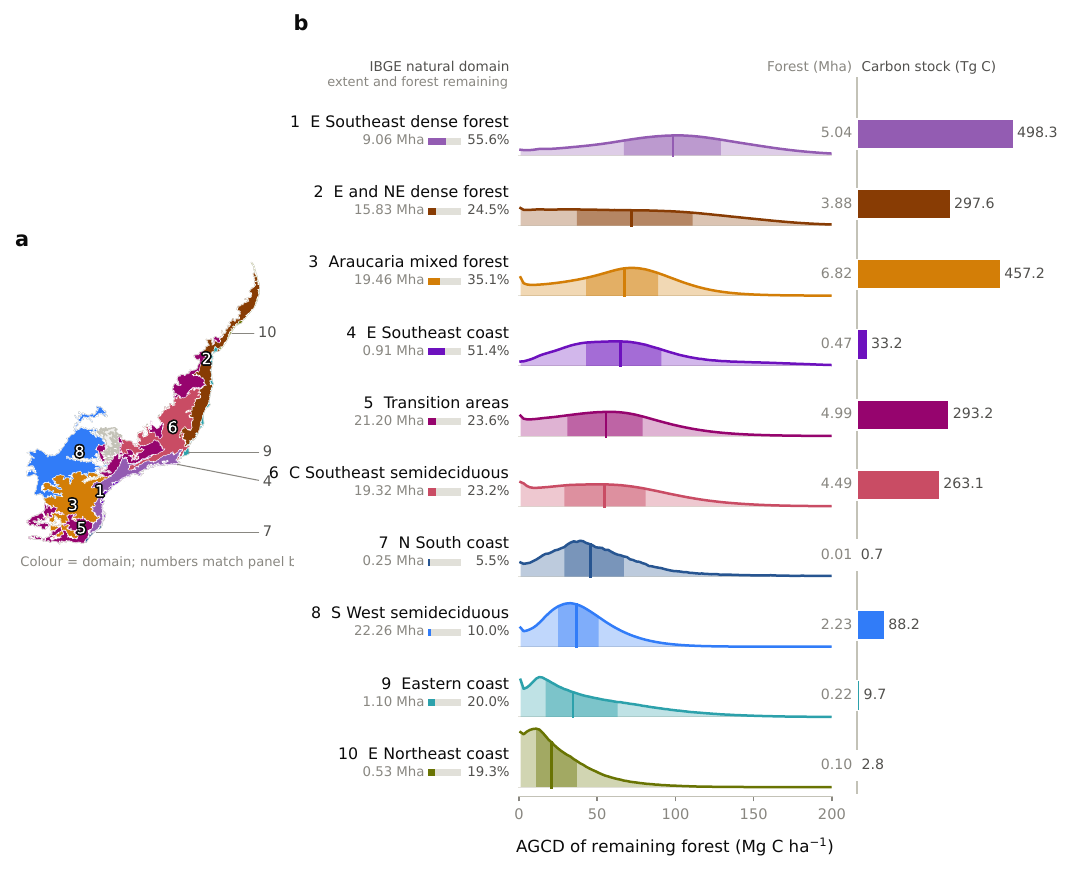}
\caption{Aboveground carbon density of the remaining forest in the ten natural domains of the Atlantic Forest. (a) The domains coloured by their mean AGCD, with the part of the region that the domain layer does not cover shown in grey. (b) Area-weighted distribution of the 30~m AGCD inside each domain, in classes of 2~Mg~C~ha$^{-1}$ and ordered by median, with the interquartile range in the darker shade and the median as a vertical line, followed by the forest area of the domain and by its total carbon stock. Colour carries the mean AGCD on the same scale in both panels, and the numbers on the map follow the order of the bands.}
\label{Fig:dominios}
\end{figure}

\section*{Usage Notes}

\subsection*{Choosing the resolution}

The canopy height layer is delivered at the native 10~m resolution of the Sentinel-2 grid because that is the scale at which the boundary-preserving branch was designed to work and at which the edges of fragments are resolved. Users whose quantity of interest is a stock over an area rather than the height of a single pixel should aggregate before using the values. About 16\% of the pixel-level error at 10~m is spatially uncorrelated and cancels inside a 30~m cell, which lowers the root mean square error from 3.05 to 2.58~m without changing the bias (``Accuracy across sensor configurations''). The carbon layer is distributed at 30~m to match the size of the calibration field plots. Further aggregation to 90~m or 1~km reduces the random component further while leaving the systematic component.

\subsection*{Reading the height-dependent bias}

The pooled bias of the model is close to zero, but it is the residue of two opposite errors. Short canopies below 10~m are overestimated by about 1.3~m and tall canopies above 30~m are underestimated by 6.5~m in the 30 to 40~m class and 16~m above 40~m (Fig.~\ref{Fig:biasByHeightAnnual}). The bias of the published product also grows with height at the three external sites, from an overestimation of 2.1 to 3.5~m below 10~m to an underestimation of 6.5~m between 30 and 40~m at Serra do Conduru and of 7.4~m above 40~m at Rio Claro (Table~\ref{tab:bias_by_height_all}). Users should therefore expect the height layer to compress the upper tail of the canopy height distribution, and the carbon layer to inherit that compression, amplified by the exponent of Equation~\ref{eq:agcd}, which is above unity. For analyses in which the tallest stands matter disproportionately, such as the stock of old-growth remnants on the coastal escarpment, the values in the upper classes should be treated as lower bounds. The bias by height class (Fig.~\ref{Fig:biasByHeightAnnual} and Table~\ref{tab:bias_by_height_all}) documents this compression. It is stratified by the reference height measured by LiDAR, which users of the map do not have. Because the model compresses the upper tail, pixels whose true height is above 30~m are mapped mostly into lower classes, so the class biases cannot be applied directly to classes of predicted height, where they would be misallocated and would create steps at the class limits. A correction requires a calibration of reference height against predicted height, which can be derived wherever airborne LiDAR is available.

\subsection*{Extrapolation outside the calibration window}

The network was trained only on the S\~ao Paulo patches, and the three external surveys share no pixel with them. The external validation therefore shows how the product behaves away from the calibration data along different axes, stature in Bahia, relief and latitude in Santa Catarina, and very tall canopy with forest plantations at Rio Claro, and the results are heterogeneous. At the tall lowland rainforest of Serra do Conduru the pooled bias is +0.33~m and the MAE is 2.82~m, against +0.19 and 2.26~m in the test patches. At the escarpment forest of Jaragu\'a do Sul the product overestimates the canopy by 2.11~m, mostly in the classes below 20~m. At Rio Claro, where half of the canopy stands above 30~m, the MAE rises to 5.47~m while the pooled bias stays at +1.46~m. The LiDAR of Conduru and Jaragu\'a do Sul precedes the product by nine and seven years, so the errors measured there include the change of the forest in the interval. Users working in the northern half of the Atlantic Forest, where the only LiDAR available to us lies in southern Bahia, should regard the Bahia figures as the most relevant indication of performance and should read the error there as a conservative bound.

\subsection*{What the carbon layer does and does not represent}

Equation~\ref{eq:agcd} converts a canopy height into the carbon density of an average stand of that height across the four categories of plot used to fit it. It does not know whether a given pixel is an old remnant, a degraded one, a natural regrowth or a restoration plantation. The calibration shows that for the same height a conserved remnant stores 1.25~kg~m$^{-2}$ more carbon than the curve predicts and a restoration plantation 0.36~kg~m$^{-2}$ less (Fig.~\ref{Fig:agcdModel}). Users who have ancillary information on forest condition, for instance from forest age products \citep{silvajunior2020benchmark} or from inventories, can improve the estimate by shifting the curve by these amounts. Without such information, the carbon layer should be read as an estimate of the stock that a stand of the mapped height typically holds, and the structural residual should be included in any uncertainty budget alongside the height error and the allometric error of the plots. The standard error of ``Uncertainty of the carbon density map'' already carries the structural residual and the random part of the allometric error through its residual term, so only the error of the predicted height remains to be added to it, with the first-order term of Equation~\ref{eq:var_height}.

\subsection*{Temporal reference and the forest mask}

Both layers describe the year 2024, the year of the annual composites. The LiDAR used for training was acquired in 2023 and 2024, so the training target and the predictors are contemporaneous to within one growing season. The external validation, in contrast, compares the 2024 product with LiDAR flown in 2015, 2017 and 2026 (``External validation sites''). The carbon layer is masked to the MapBiomas 2024 forest classes, whose pixels are those with valid values in record 1. Because the mask is itself a classification with its own commission and omission errors, the carbon stock of a region depends on the mask as much as on the height model, and the two sources of uncertainty should be reported separately. Records 5 and 6, distributed without the mask, make it possible to recompute height and carbon stock under any alternative definition of forest.

\subsection*{Intended uses}

The dataset was built for three kinds of applications. The first is carbon accounting at regional scale and for the Atlantic Forest as a whole, where the aggregation to 30~m and above gives a product whose random error is small and whose systematic error is documented by height class. The second is the study of forest dynamics at the scale of the fragment, where the 10~m resolution resolves edges, corridors and small remnants that coarser products merge with the matrix, and where the combination with annual land cover and forest age layers allows carbon to be tracked along chronosequences of regrowth. The third is the prioritization of restoration, for which the map provides the baseline stock and, when combined with the age of regenerating fragments, the realized rate of accumulation across the environmental gradients of the region. In all three cases, the class-wise bias and the standard errors of records 3 and 4 should be propagated into any derived quantity, and the training footprints, in S\~ao Paulo, should be excluded when the product is evaluated against new reference data.

Taken together, the two layers provide the first height and carbon reference for the Atlantic Forest that was calibrated inside the region, validated outside its calibration window and documented pixel by pixel against airborne LiDAR. The comparison with the global 1~m product over the same pixels shows that a model trained on regional LiDAR with multi-sensor fusion lowers the mean absolute error of a model trained to work everywhere by 46\% over the test patches and by 22 to 59\% at three independent sites, and the carbon accounting by natural domain shows that the map reproduces the structural gradient of the region. The remaining limitations, the compression of tall canopies, the structural residual of the height-to-carbon relationship, and the heterogeneity of performance across the external sites, are quantified rather than removed, and the records needed to correct for them are part of the dataset.

\section*{Code availability}

The code that trains the multi-sensor network and applies it to new patches is available at \url{https://github.com/mpferreiraUSP/AtlanticForest_CHM}. The repository contains three Python scripts and their documentation. \texttt{model.py} defines the network, with one branch per sensor fed at its native resolution, the gated fusion of the branch features on the Sentinel-2 grid and the regression head, together with the normalization of the inputs and the loading of checkpoints. \texttt{train.py} pairs the patches of each sensor by file name, computes the normalization statistics, trains the network with the mean absolute error and early stopping on the validation error, and writes checkpoints that carry the architecture and the normalization constants. A branch is created only when the folder of its sensor is given, so the same script trains every sensor configuration of ``Experimental design''. \texttt{predict.py} rebuilds the network from a checkpoint, normalizes the inputs as in training, and writes one canopy height raster per Sentinel-2 patch, georeferenced when the inputs are GeoTIFF files. The file \texttt{requirements.txt} lists the dependencies, namely PyTorch 2.0 or later, \texttt{segmentation-models-pytorch} 0.3 or later, NumPy, rasterio, Matplotlib and tqdm, and the README describes the expected layout of the patches and the training options. 

\bibliography{biblio}

\section*{Acknowledgements}
We thank the \emph{Instituto Geogr\'afico e Cartogr\'afico do Estado de S\~ao Paulo} (IGC-SP) for the statewide airborne LiDAR campaign, and the Sustainable Landscapes Brazil project for the airborne LiDAR surveys of Serra do Conduru and Jaragu\'a do Sul. This study was funded by Fundação de Amparo à Pesquisa do Estado de São Paulo (FAPESP), grants \#2024/15211-1, \#2022/15507-2, \#2023/09862-7, \#2018/18416-2, \#2020/06734-0, \#2022/14605-0, \#2019/24049-5, \#2021/10573-4, \#2014/50279-4 and \#2020/15230-5. M.P.F. thanks Conselho Nacional de Desenvolvimento Científico e Tecnológico (CNPq), grant \#309000/2023-8. Additional support came from the Dutch Research Council (grant \#5160957745), Shell Brazil (grant \#22047-5), AES Brasil (grant \#104403), WeForest (grant \#102946), Instituto de Pesquisas Ecológicas (grant \#103775), The Nature Conservancy (grant \#104004), World Wide Fund for Nature (grant \#103974), Brascell, Re.green (grant \#104424), Suzano (grant \#103908) and Sylvamo (grant \#104251).

\section*{Author contributions}
M.P.F. conceived the study, designed the network and experiments, processed the LiDAR and satellite data, produced the maps, and wrote the original draft. T.C.R. provided access to the LiDAR data. G.B.S. compiled the field inventory data and plot-level LiDAR metrics, and contributed to the carbon calibration. G.A.F. and R.C.P. processed the LiDAR data. D.R.A.A. and P.G.M. co-conceived and coordinated the study, and contributed to writing the manuscript. C.T.A., and A.F.R. compiled the field data and participated in the LiDAR data acquisition. V.C.S., M.A.P.F. and C.T. performed the botanical identification of tree species from the field plots. E.N.B and A.A.Z. performed LiDAR data acquisition over the field plots. J.v.M. compiled and curated the field data. L.S. participated in field investigation and field data curation. R.R.R supervised the project. P.H.S.B. supervised the project and acquired funding and edited the original draft. All authors reviewed and approved the final manuscript.

\section*{Competing interests}
M.P.F., D.R.A.A., and P.G.M. are partners at Bioflore, a forest monitoring technology company. P.H.S.B. is a partner at Re.green, a forest restoration company.

\clearpage
\section*{Supplementary Information}
\label{sec:supplementary}
\setcounter{table}{0}
\renewcommand{\thetable}{S\arabic{table}}
\providecommand{\theHtable}{}
\renewcommand{\theHtable}{S\arabic{table}}

\begin{table}[H]
\centering
\caption{Accuracy of canopy height predicted from the Sentinel-2 annual composite alone (experiment E1) by three encoder--decoder architectures, evaluated against airborne LiDAR reference height within the MapBiomas forest mask on the S\~ao Paulo test patches.}
\label{tab:arch_comparison_E1_p10}
\begin{adjustbox}{max width=\textwidth}
\begin{threeparttable}
\begin{tabular}{llrrrrrrr}
\toprule
Architecture & Encoder & Params (M) & RMSE (m) & MAE (m) & Bias (m) & Bias (\%) & $R^2$ & CCC \\
\midrule
BPAT-UNet & -- & 43.6 & \textbf{3.35} & \textbf{2.50} & +0.17 & +1.1 & \textbf{0.714} & \textbf{0.835} \\
U-Net & ResNet-50 (ImageNet) & 36.4 & 3.47 & 2.60 & \textbf{$-$0.06} & $-$0.4 & 0.694 & 0.825 \\
MA-Net & ResNet-50 (ImageNet) & 64.9 & 3.38 & 2.52 & +0.26 & +1.6 & 0.711 & \textbf{0.835} \\
\bottomrule
\end{tabular}
\begin{tablenotes}
\footnotesize
\item All networks were trained with the same data (S\~ao Paulo training patches and the same training and validation split), input normalization, loss function (MAE), optimizer, learning-rate schedule and early stopping (patience of 10 epochs on the validation MAE), and only the architecture differs. U-Net and MA-Net use a ResNet-50 encoder initialized with ImageNet weights, whereas BPAT-UNet has no pretrained encoder and is trained from random initialization. Metrics computed on $n = 55{,}130{,}381$ forest pixels per model, inside the MapBiomas forest mask and with LiDAR height $> 0$~m. Bias~=~mean(LiDAR $-$ predicted), and Bias~\% is relative to the mean reference height. $R^2$ from the ordinary least-squares fit of predicted vs.\ reference height. CCC is Lin's concordance correlation coefficient and Params the number of trainable parameters. Bold marks the best value per column at the precision shown.
\end{tablenotes}
\end{threeparttable}
\end{adjustbox}
\end{table}

\end{document}